\documentclass{article}

\usepackage{microtype}
\usepackage{graphicx}
\usepackage{booktabs} %
\usepackage{multirow}
\usepackage{enumitem}
\usepackage{graphicx}
\usepackage{caption}
\usepackage{subcaption}
\usepackage[makeroom]{cancel}

\usepackage{thm-restate}

\usepackage[accepted]{style/icml2025}

\usepackage{amssymb}%
\usepackage{pifont}
\usepackage{amsmath}
\usepackage{amssymb}
\usepackage{mathtools}
\usepackage{amsthm}
\usepackage{xspace}

\usepackage{siunitx}

\usepackage[final, colorlinks=true, allcolors=blue, pagebackref=true]{hyperref}
\usepackage[capitalize,noabbrev]{cleveref}

\usepackage{amsthm}
\usepackage{xcolor}
\usepackage{amsmath,amsfonts,bm}
\makeatletter
\newtheorem*{rep@theorem}{\rep@title}
\newcommand{\newreptheorem}[2]{%
\newenvironment{rep#1}[1]{%
 \def\rep@title{#2 \ref{##1}}%
 \begin{rep@theorem}}%
 {\end{rep@theorem}}}
\makeatother

\newtheorem{proposition}{Proposition}

\newreptheorem{theorem}{Theorem}
\newtheorem{lemma}{Lemma}

\definecolor{myred}{RGB}{215,48,39}
\definecolor{mygreen}{RGB}{26,152,80}
\newcommand{\cmark}{\textcolor{mygreen}{\ding{51}}}
\newcommand{\xmark}{\textcolor{myred}{\ding{55}}}
\newcommand{\halfmark}{\textcolor{gray}{\checkmark\kern-1.1ex\raisebox{.7ex}{\rotatebox[origin=c]{125}{--}}}}
\usepackage[framemethod=TikZ]{mdframed}
\mdfdefinestyle{MyFrame}{%
    linecolor=black,
    outerlinewidth=.3pt,
    roundcorner=5pt,
    innertopmargin=1pt, %
    innerbottommargin=1pt, %
    innerrightmargin=1pt,
    innerleftmargin=1pt,
    backgroundcolor=black!0!white}

\mdfdefinestyle{MyFrame2}{%
    linecolor=white,
    outerlinewidth=1pt,
    roundcorner=1pt,
    innertopmargin=3pt,
    innerbottommargin=2pt,
    innerrightmargin=7pt,
    innerleftmargin=7pt,
    backgroundcolor=black!3!white}

\mdfdefinestyle{MyFrameEq}{%
    linecolor=white,
    outerlinewidth=0pt,
    roundcorner=0pt,
    innertopmargin=0pt,
    innerbottommargin=0pt,
    innerrightmargin=7pt,
    innerleftmargin=7pt,
    backgroundcolor=black!3!white}

\newcommand{\RNum}[1]{\uppercase\expandafter{\romannumeral #1\relax}}

\newcommand{\E}{\mathcal{E}}

\newcommand{\R}{\mathcal{R}}

\newcommand{\vertiii}[1]{{\left\vert\kern-0.25ex\left\vert\kern-0.25ex\left\vert #1 
    \right\vert\kern-0.25ex\right\vert\kern-0.25ex\right\vert}}
\newcommand{\vertiiii}[1]{{\vert\kern-0.25ex\vert\kern-0.25ex\vert #1 
    \vert\kern-0.25ex\vert\kern-0.25ex\vert}}

\usepackage{mathtools}

\newcommand{\xhdr}[1]{{\noindent\bfseries #1}.}
\newcommand{\cut}[1]{}

\newcommand{\removelatexerror}{\let\@latex@error\@gobble}

\def\eqref#1{Eq.~\ref{#1}}

\def\1{\bm{1}}

\def\eps{{\epsilon}}

\def\vx{{\bm{x}}}

\DeclareMathAlphabet{\mathsfit}{\encodingdefault}{\sfdefault}{m}{sl}
\SetMathAlphabet{\mathsfit}{bold}{\encodingdefault}{\sfdefault}{bx}{n}

\def\gD{{\mathcal{D}}}
\def\gE{{\mathcal{E}}}

\def\gL{{\mathcal{L}}}

\def\gN{{\mathcal{N}}}

\def\gU{{\mathcal{U}}}

\def\gW{{\mathcal{W}}}

\def\gZ{{\mathcal{Z}}}

\def\sE{{\mathbb{E}}}

\def\sP{{\mathbb{P}}}

\def\R{{\mathbb{R}}}

\def\sT{{\mathbb{T}}}

\newcommand{\KL}{\mathbb{D}_{\mathrm{KL}}}

\newcommand{\sethree}{\mathrm{SE(3)}}

\newcommand{\sothree}{\mathrm{SO(3)}}

\newcommand{\namelong}{\textsc{Sequential Boltzmann Generators}\xspace}
\newcommand{\nameshort}{\textsc{SBG}\xspace}

\newcommand{\emetric}{\(\mathcal{E}\text{‑}\mathcal{W}_2\)\xspace}
\newcommand{\torusmetric}{\(\mathbb{T}\text{‑}\mathcal{W}_2\)\xspace}
\newcommand{\ticametric}{\(\mathrm{TICA}\text{‑}\mathcal{W}_2\)\xspace}

\newcommand{\sci}[2]{\(#1\cdot10^{#2}\)}
\newcommand{\scione}[1]{\(10^{#1}\)}

\makeatletter
\renewcommand*{\appendixautorefname}{\S\@gobble}
\renewcommand*{\sectionautorefname}{\S\@gobble}
\renewcommand*{\subsectionautorefname}{\S\@gobble}
\renewcommand*{\subsubsectionautorefname}{\S\@gobble}
\makeatother

\usepackage[createShortEnv,conf={no link to proof, text link},commandRef=autoref]{proof-at-the-end}

\renewcommand*{\backrefalt}[4]{%
    \ifcase #1 \footnotesize{(Not cited.)}%
    \or        \footnotesize{(Cited on page~#2)}%
    \else      \footnotesize{(Cited on pages~#2)}%
    \fi}

\hypersetup{
	colorlinks=true,       %
	linkcolor=blue,        %
	citecolor=blue,        %
	filecolor=magenta,     %
	urlcolor=blue         
}

\icmltitlerunning{Scalable Equilibrium Sampling with Sequential Boltzmann Generators}

\begin{document}

\twocolumn[
\icmltitle{Scalable Equilibrium Sampling with Sequential Boltzmann Generators}

\icmlsetsymbol{equal}{*}

\begin{icmlauthorlist} 
\icmlauthor{Charlie B. Tan}{equal,oxford}
\icmlauthor{Avishek Joey Bose}{equal,oxford,mila}
\icmlauthor{Chen Lin}{oxford}
\icmlauthor{Leon Klein}{berlin}
\icmlauthor{Michael M. Bronstein}{oxford,aithyra}
\icmlauthor{Alexander Tong}{mila,udem}
\end{icmlauthorlist}

\icmlaffiliation{mila}{Mila -- Qu\'ebec AI Institute}
\icmlaffiliation{udem}{Universit\'e de Montr\'eal}
\icmlaffiliation{oxford}{University of Oxford}
\icmlaffiliation{berlin}{Freie Universität Berlin}
\icmlaffiliation{aithyra}{AITHYRA}

\icmlcorrespondingauthor{CBT}{charlie.tan@exeter.ox.ac.uk}
\icmlcorrespondingauthor{AJB}{joey.bose@mail.mcgill.ca}
\icmlcorrespondingauthor{AT}{alexandertongdev@gmail.com}

\icmlkeywords{Machine Learning, Boltzmann Generators, Importance Sampling, Molecules, Normalizing Flows}

\vskip 0.3in
]

\printAffiliationsAndNotice{\icmlEqualContribution} %

\begin{abstract}

Scalable sampling of molecular states in thermodynamic equilibrium is a long-standing challenge in statistical physics. Boltzmann generators tackle this problem by pairing normalizing flows with importance sampling to obtain uncorrelated samples under the target distribution. In this paper, we extend the Boltzmann generator framework with two key contributions, denoting our framework \namelong (\nameshort). The first is a highly efficient Transformer-based normalizing flow operating directly on all-atom Cartesian coordinates. In contrast to the equivariant continuous flows of prior methods, we leverage exactly invertible non-equivariant architectures which are highly efficient during both sample generation and likelihood evaluation. This efficiency unlocks more sophisticated inference strategies beyond standard importance sampling. In particular, we perform inference-time scaling of flow samples using a continuous-time variant of sequential Monte Carlo, in which flow samples are transported towards the target distribution with annealed Langevin dynamics. \nameshort achieves state-of-the-art performance w.r.t.\ all metrics on peptide systems, demonstrating the first equilibrium sampling in Cartesian coordinates of tri-, tetra- and hexa-peptides that were thus far intractable for prior Boltzmann generators. 

\end{abstract}

\section{Introduction}
\label{sec:introduction}

\begin{figure}[t]
    \includegraphics[width=\linewidth]{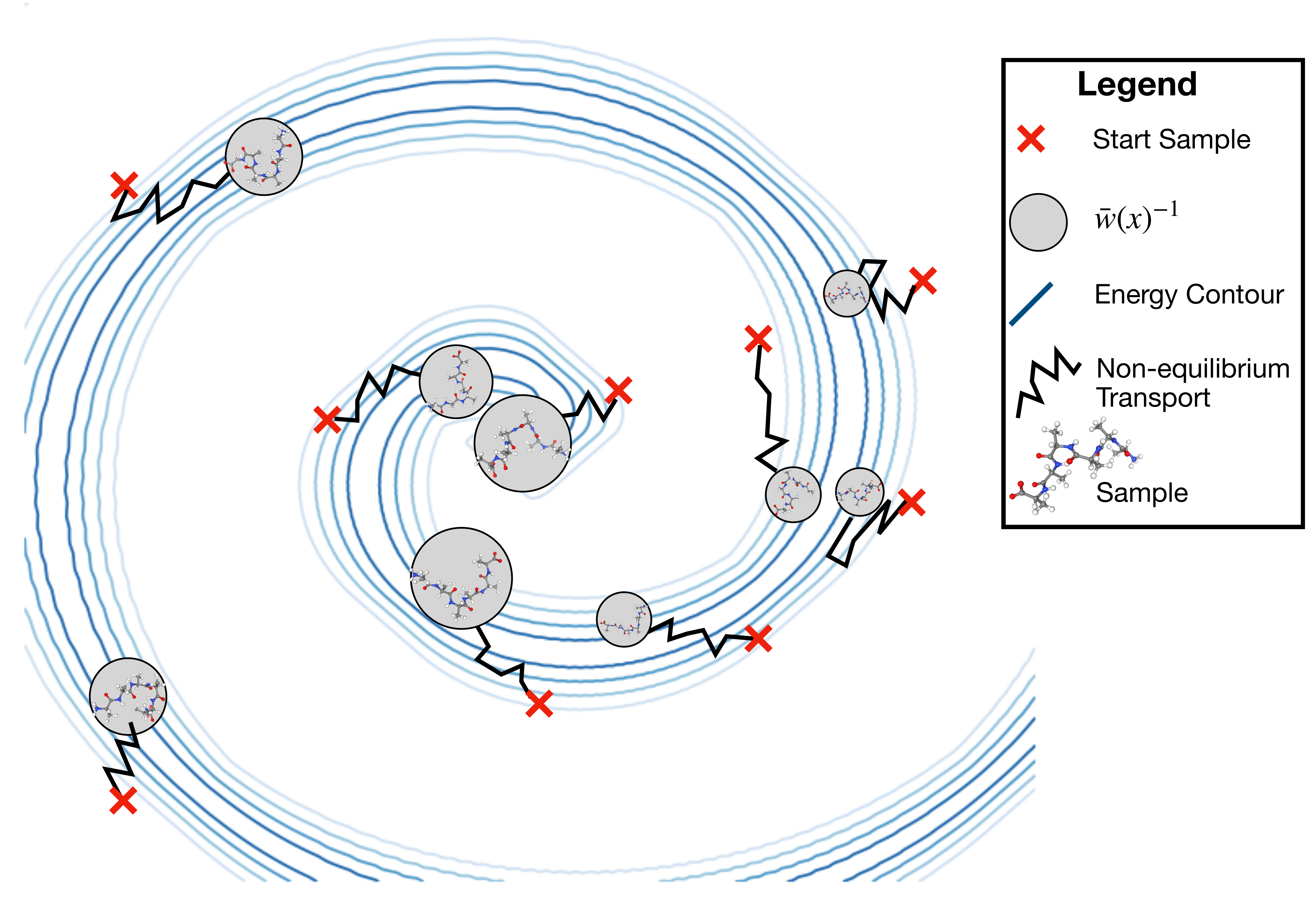}
    \caption{
        \small \nameshort uses annealed Langevin dynamics to transport proposal flow samples towards towards the target distribution.
    }
    \vspace{-8pt}
    \label{fig:visual_abstract}
\end{figure}

The sampling of molecular systems at the all-atom resolution is of central interest in understanding complex natural processes. These include important biophysical processes such as protein-folding~\citep{noe2009constructing,lindorff2011fast}, protein-ligand binding~\citep{buch2011complete}, and formation of crystal structures~\citep{parrinello1980crystal,matsumoto2002molecular}, whose understanding can aid in problems that range from long-standing global health challenges, to efficient energy storage~\citep{Deringer_2020}. 

The dominant paradigm for molecular sampling involves running Markov chain Monte Carlo (MCMC) or molecular dynamics (MD), whereby the equations of motion are integrated with finely discretized time steps. However, such molecular systems often exist in thermodynamic equilibrium by remaining for extended periods in metastable states. Such metastable states are captured in the minima of a complex energy landscape, itself defining the molecular system's equilibrium (Boltzmann) distribution at a given temperature. The high-energy barriers separating metastable states lead to infrequent state transitions \citep{wirnsberger2020targeted}, presenting an obstacle for effective sampling with simulation-based methods such as molecular dynamics or MCMC, requiring long simulation periods with small time steps on the order of femtoseconds \SI{1}{\fs} $=$ \scione{-15}\SI{}{\s}.

Boltzmann generators (BG)~\citep{noe2019boltzmann} offer an alternative approach, in which powerful generative models, such as normalizing flows~\citep{dinh2016density,rezende2015variational}, are trained on existing (but assumed to be biased) datasets, and leveraged as a proposal for self-normalized importance sampling (SNIS), targeting the desired Boltzmann distribution. Boltzmann generators permit accelerated sampling through amortization as the uncorrelated proposal generation avoids the slow state transitions suffered by MD and MCMC. Despite their appeal, it remains challenging for existing BGs to model systems beyond the smallest peptides (2 amino acids) in Cartesian coordinates~\citep{klein2023equivariant,midgley2023se}. The principal drawback inhibiting scalability stems from the lack of expressive equivariant architectures that are also exactly invertible~\citep{bose2021equivariant,midgley2023se}, or the present over-reliance on simple $\text{E}(n)$-GNN~\citep{satorras2021n} based equivariant vector fields in continuous-time normalizing flows~\citep{chen_neural_2018}. As a result, even the most performant BGs suffer from poor target distribution overlap, leading to low sampling efficiency during SNIS.

\xhdr{Present work}
In this paper, we introduce \namelong (\nameshort) a novel extension to the existing Boltzmann generator framework.\footnote{We open source our full codebase at\\ \url{https://github.com/transferable-samplers/transferable-samplers}.} \nameshort makes progress on the scalability of Boltzmann generators in Cartesian coordinates along two complementary axes:  (1) scalable pre-training of softly $\sethree$-equivariant proposal normalizing flows in BGs;  and (2) inference time scaling via continuous-time variants of annealed importance sampling (AIS)~\citep{neal2001annealed} and sequential Monte Carlo (SMC)~\citep{doucet2001sequential}. The use of AIS or SMC over SNIS enables more effective sampling given suboptimal proposal-target overlap, enabling \nameshort to draw uncorrelated \(\mu_\text{target}(x)\) samples for peptide system up to 6 residues.

\begin{table}[htb]
    \vspace{-5pt}
    \centering
    \caption{\small Method overview for samplers, given biased data samples.}
\resizebox{1\columnwidth}{!}{
\begin{tabular}{l|cccccc}
\toprule
Method  & Use $\mathcal{E}(x)$ & Exact likelihoods & Use data & Annealing\\
\midrule
DEM~\citep{akhound2024iterated}                 & \cmark & \xmark & \xmark & \xmark \\
NETS~\citep{albergo_nets_2024}                  & \cmark & \cmark & \xmark & \cmark \\
BG~\citep{noe2019boltzmann}                 & \cmark & \cmark & \cmark & \xmark \\
\nameshort (Ours) & \cmark & \cmark & \cmark & \cmark\\
\bottomrule
\end{tabular}
}
\label{tab:summary}
\end{table}

\nameshort scales up normalizing flows in BGs by following recent advances in atomistic generative modeling \citep{abramson2024accurate}. In particular, we remove the rigid $\sethree$-equivariance as an explicit architectural inductive bias in favor of softly enforcing it through simpler and more efficient data augmentations. To further improve sampling  we perform inference-time scaling by defining an interpolation between the proposal flow energy distribution (i.e., negative log density of samples) and the known target Boltzmann energy. Crucially, simulating samples at inference via annealed Langevin dynamics may be coupled to a corresponding time evolution of importance weights, converting naturally to continuous-time variants of the well-established annealed importance sampling (AIS)~\citep{neal2001annealed} and sequential Monte Carlo (SMC)~\citep{doucet2001sequential}. As a result, \nameshort can readily improve over the simple one-step importance sampling methodology used in existing BGs. We summarize the different aspects of our proposed \nameshort in comparison to other learned samplers in~\cref{tab:summary}.

We instantiate \nameshort using a best-in-class, general-purpose, \emph{non-equivariant} normalizing flow, named TarFlow~\citep{zhai2024normalizing}. TarFlow is a modernized normalizing flow architecture employing a scalable transformer backbone to parameterize an exactly invertible transformation. We demonstrate that such exactly invertible architectures, via fast and accurate log-likelihood evaluation, benefit from inference-scaling. We emphasize this is in stark contrast to continuous normalizing flows that underpin prior SOTA Boltzmann generators which require both the costly simulation of the 2nd order divergence operator as well as differentiation of an ODE solver. Furthermore, we demonstrate that enforcing equivariance softly along enables us to stably scale proposal flows in \nameshort, far beyond prior BGs. On a theoretical front, we study a novel inference-time proposal energy adjustment to counteract the influence of training data centroid augmentation when resampling, as well as quantify the additional bias of common thresholding tricks employed to improve resampling numerical stability. Empirically, we observe \nameshort to achieve state-of-the-art results across metrics, far outperforming continuous BGs on all datasets. In particular, \nameshort is the first uncorrelated learned sampler to scale successfully in Cartesian coordinates to tripeptides, tetrapeptides, hexapeptides, and makes significant progress towards equilibrium sampling of decapeptides.

\section{Background and Preliminaries}
\label{sec:background}

We are interested in drawing statistically independent samples from the target Boltzmann distribution $\mu_{\text{target}}$, with partition function $\gZ$, defined over $\R^{n \times 3}$:
\begin{equation*}
    \mu_{\text{target}}(x) \propto \exp\left( \frac{-\E(x)}{k_{\text{B}}T} \right), \gZ = \int_{\mathbb{R}^d} \exp\left( \frac{-\E(x)}{k_{\text{B}}T} \right) dx.
\end{equation*}

The Boltzmann distribution is defined for a given system and includes the Boltzmann constant $k_{\text{B}}$, and a specified temperature $T$. Additionally, the potential energy of the system $\gE: \R^{n \times 3} \to \R$ and its gradient $\nabla \gE$ can be evaluated at any point $x \in \R^{n \times 3}$, but the exact density $\mu_{\text{target}}(x)$ is not available as the partition function $\gZ$ evaluation is intractable for all but the simplest systems.

In this paper, unlike pure sampling-based settings, we are afforded access to a small biased dataset of $N$ samples $\gD = \{ x^i \}_{i=1}^{N}$, provided as an empirical distribution $p_{\gD}$. Consequently, it is possible to perform an initial learning phase that fits a generative model $p_{\theta}$, with parameters $\theta$, to $p_{\gD}$---e.g. by minimizing the forward KL $\KL(p_{\gD} || p_{\theta})$---to act as a proposal distribution that can be corrected. 

\subsection{Normalizing Flows}
\label{sec:normalizing_flows}

A key desirable property needed for the correction of a trained generative model $p_{\theta}$ on a biased dataset $\gD$ is the ability to extract an exact likelihood $p_{\theta}(x)$. Normalizing flows~\citep{dinh2016density,rezende2015variational} represent exactly such a model class as they learn to transform an easy-to-sample base density to a desired target density using a parametrized diffeomorphism. More formally, given a sample from a (prior) base density $x_0 \sim p_0$ and a diffeomorphism $f_{\theta}: \R^{n \times 3} \to \R^{n \times 3}$ that maps the initial sample to $x_1 = f_{\theta}(x_0)$. We can obtain an expression for the log density of $x_1$ via the classical change of variables, %
\begin{equation}
    \log p_1(x_1) = \log p_0(x_0) - \log \det \left | \frac{\partial f_{\theta} (x_0)}{\partial x_0}\right |.
    \label{eqn:change_of_variables}
\end{equation}
\looseness=-1
In~\eqref{eqn:change_of_variables} above the $\log \det | \cdot |$ term corresponds to the Jacobian determinant of $f_{\theta}$ evaluated at $x_0$. 
Optimizing \eqref{eqn:change_of_variables} is the maximum likelihood objective for training normalizing flows and results in $f_{\theta}$ learning $p_1 \approx p_{\text{data}}$.
There are multiple ways to construct the (flow) map $f_{\theta}$. Perhaps the most popular approach is to consider the flow to be a composition of a finite number of elementary diffeomorphisms $f_{\theta} = f_M \circ f_{M-1} \dots \circ f_1$, resulting in the change in log density to be: $ \log p_1(x_1) = \log p_0(x_0) - \sum^M_{i=1} \log \left | \partial f_{i, \theta} (x_{i-1})/ \partial x_{i-1} \right |$. We note that the construction of each $f_{i, \theta},  i\in [M]$ is motivated such that both the inverse $f^{-1}_{i, \theta}(x)$ and Jacobian $\partial f_{i, \theta} (x) / \partial x $ are computationally cheap to compute.

\xhdr{Continuous normalizing flows}
In the limit of infinite elementary diffeomorphisms, a normalizing flow transforms into a continuous normalizing flow (CNF)~\citep{chen_neural_2018}. Formally, a \emph{flow} is a one-parameter time-dependent diffeomorphism $\psi_{t}:[0,1] \times \R^{n \times 3} \to \R^{n \times 3}$ that is the solution to the following ordinary differential equation (ODE): $ \frac{d}{dt} \psi_{t}(x) = u_t \left(\psi_{t}(x) \right)$, with initial conditions $\psi_0(x_0) = x_0$, for a time-dependent vector field $u_t:[0,1]\times\R^{n \times 3}\to\R^{n \times 3}$. It is often desirable to construct the target flow by associating it to a designated \emph{probability path} $p_t: [0, 1] \times \sP(\R^{n \times 3}) \to \sP(\R^{n \times 3})$ which is a time-indexed interpolation in probability space between two distributions $p_0, p_1 \in \sP(\R^{n \times 3})$. 
In such cases, the flow $\psi_t$ is said to generate $p_t$ if it pushes forward $p_0$ to $p_1$ by following $u_t$ --- $p_t = [\psi_t]_\# (p_0)$.
As $\psi_t$ is a valid flow and satisfies an ODE the change in log density can be computed using the instantaneous change of variables:
\begin{equation}
    \log p (x_1) = \log p(x_0) - \int^1_0 \nabla \cdot u_t (x_t) dt,
    \label{eqn:instantaneous_change_of_variable}
\end{equation}
where $x_t = \psi_t(x_0)$ and $\nabla \cdot$ is the divergence operator. 

A CNF can then be viewed as a neural flow that seeks to learn a designated target flow $\psi_t$ for all time $t\in [0,1]$. 
The most scalable way to train CNFs is to employ a flow-matching learning framework~\citep{liu_rectified_2022,albergo_building_2023,lipman_flow_2022,tong_conditional_2023}. Specifically, flow-matching regresses a learnable vector field of a CNF $f_{t, \theta}(t, \cdot): [0,1] \times \R^{n \times 3} \to \R^{n \times 3}$ to the target vector field $u_t(x_t)$ associated to the flow $\psi_t$. In practice, it is considerably easier to regress against a target \emph{conditional} vector field $u_t (x_t | z)$---which generates the conditional probability path $p_t(x_t|z)$---as we do not have closed form access to the (marginal) vector field $u_t$ which generates $p_t$. The conditional flow-matching (CFM) objective can then be stated as a simple simulation-free regression,
\begin{equation}
\gL_{\rm CFM}(\theta) = \mathbb{E}_{t, q(z), p_t(x_t | z)} \|f_{t,\theta}(t, x_t) - u_t(x_t | z)\|_2^2.
\label{eqn:CFM}
\end{equation}
The conditioning distribution $q(z)$ can be chosen from any valid coupling, for instance, the independent coupling $q(z)= p(x_0) p(x_1)$. 
We highlight that~\eqref{eqn:CFM} allows for greater flexibility in $f_{t, \theta}$ as there is no exact invertibility constraint. To generate samples and their corresponding log density according to the CNF we may solve the following flow ODE numerically with initial conditions $x_0 = \psi_0(x_0)$ and $c = \log p_0 (x_0)$, which is the log density under the prior: 
\begin{equation}
    \frac{d}{dt} 
    \begin{bmatrix}
        \psi_{t, \theta}(x_t) \\
        \log p_t (x_t)
    \end{bmatrix} = 
    \begin{bmatrix}
        f_{t, \theta}(t, x_t) \\
        -\nabla \cdot f_{t, \theta}(t, x_t)
    \end{bmatrix}.
    \label{eqn:cnf_and_log_prob_ode}
\end{equation}

\subsection{Boltzmann Generators}
\label{sec:boltzmann_generators}

A Boltzmann generator~\citep{noe2019boltzmann} $\mu_{\theta}$ pairs a normalizing flow as the proposal generative model $p_{\theta}$, which is then corrected to obtain i.i.d. samples under $\mu_{\text{target}}$ using self-normalized importance sampling. More precisely, as normalizing flows are exact likelihood models, BG's first draw $K$ independent samples $x^i \sim p_{\theta}(x), i\in [K]$ and compute the corresponding (unnormalized) importance weights for each sample $w(x^i) = \exp\left(\frac{-\gE(x^i)}{k_\text{BT}} \right)/ p_{\theta}(x^i)$. Leveraging the importance weights we can compute a Monte-Carlo approximation to any observable $\phi(x)$ of interest under $\mu_{\text{target}}$ using self-normalized importance sampling as follows:
\begin{equation*}
     \mathbb{E}_{\mu_{\text{target}}(x)}[ \phi(x)] = \mathbb{E}_{p_\theta} [\phi(x) \bar{w}(x)] \approx \frac{\sum_{i=1}^K w(x^i) \phi(x^i)}{\sum_{i=1}^K w(x^i)}.
\end{equation*}
In addition, computing importance weights also enables resampling the pool of samples according to the collection of normalized importance weights $W = \{\bar{w}(x^i) \}_{i=1}^K$. 

\section{\namelong}
\label{sec:method}
We now present \nameshort, which extends and improves over classical Boltzmann generators by including an annealing process to transport proposal samples towards the target distribution. We begin by identifying the key limitation in current BGs as SNIS with a suboptimal proposal. Indeed, while the SNIS estimator is consistent, its efficacy is highly dependent on the overlap between proposal $p_{\theta}$ and target $\mu_\text{target}$, where the optimal proposal is proportional to the minimizer of the variance of $\phi(x^i)\mu_{\text{target}}(x^i)$~\citep{mcbook}. Unfortunately, since $p_{\theta}$ within a BG is trained on a biased dataset $\gD$ the importance weights typically exhibit large variance, resulting in a small effective sample size (ESS).\footnote{ESS is defined as: $\text{ESS} = 1/ \sum^K_i (\bar{w}(x^i))^2$.} 

We address the need for more flexible proposals in~\S\ref{sec:scaling_training} with modernized scalable training recipes for atomistic normalizing flows. 
In~\S\ref{sec:jarzynski} we outline our novel application of non-equilibrium sampling with sequential Monte Carlo \citep{doucet2001sequential}. We term the overall process of combining a pre-trained Boltzmann generator with inference scaling through annealing \namelong. 

\xhdr{Symmetries of molecular systems}
The energy function $\gE(x)$ in a molecular system using classical force fields is invariant under global rotations and translation, which corresponds to the group $\sethree \cong \sothree \ltimes (\R^3, +)$. Unfortunately, $\sethree$ is a non-compact group which does not allow for defining a prior density $p_0(x_0)$ on $\R^{n\times 3}$. Equivariant generative models circumvent this issue by defining a mean-free prior which is a projection of a Gaussian prior $\gN(0, I)$ onto the subspace $\R^{(n-1) \times 3}$~\citep{garcia2021n}. Thus pushing forward a mean free prior with an equivariant flow provably leads to an invariant proposal $p_1(x_1)$~\citep{kohler2020equivariant,bose2021equivariant}. We next build BGs departing from exactly equivariant maps by considering soft equivariance, unlocking scalable and efficient architectures.

\subsection{Scaling Training of Boltzmann Generators}
\label{sec:scaling_training}
To improve proposal flows in \nameshort we favor scalable architectural choices that are more expressive than exactly equivariant ones. We motivate this choice by highlighting that many classes of normalizing flow models are known to be universal density approximators~\citep{teshima2020coupling,lee2021universal}. Thus, expressive enough non-equivariant flows \emph{can learn to approximate any equivariant map}.

\xhdr{Soft equivariance}
We instantiate \nameshort with a state-of-the-art TarFlow~\citep{zhai2024normalizing} which is based on blockwise masked autoregressive flow~\citep{papamakarios2017masked} based on a causal Vision Transformer (ViT)~\citep{alexey2020image} modified for molecular systems where patches are over the particle dimension. Since the data comes mean-free we further normalize the data to unity standard deviation. Combined, this allows us to scale both the depth and width of the models stably as there is no tension between a hard equivariance constraint and the invertibility of the network.

We include a series of strategies to improve training of non-equivariant flows by softly enforcing $\sethree$-equivariance. First, we softly enforce equivariance to global rotations through data augmentation by sampling random rotations $R \in \sothree$ and applying them to data samples $R\circ x_1 \sim p_1(x_1)$. Secondly, as the data is mean-free and has $(n-1) \times 3$ degrees of freedom, we lift the data dimensionality back to $n$ by adding noise to the center of mass. This allows us to easily train with a non-translational equivariant prior distribution such as the standard normal $p_0 = \gN(0,I)$. More precisely, a data sample is constructed $x = R \bar{x} + c$, where $\bar{x} \in \R^{(n-1) \times 3} \hookrightarrow \R^{n \times 3}$ is the mean-free data point embedded in $\R^{n \times 3}$, $R \in \sothree$, and $c \sim \gN (0, \sigma^2)$. At inference, the impact of this center of mass noise is that we must account for $p(\| c\|)$, which follows a $\chi_3$ distribution in three dimensions. Consequently, during reweighting we adjust the proposal energy to account for the impact of center of mass training augmentation as follows: 
\begin{equation}
    \log p^c_{\theta}(x) = \log p_{\theta}(x) + \frac{\| c \|^2}{2\sigma^2} - \log\left[\frac{\| c^2\|}{\sqrt{2}\sigma^3\Gamma\left(\frac{3}{2}\right)}\right],
    \label{eqn:com_adjust}
\end{equation}
\looseness=-1
where \(\Gamma(\cdot)\) is the gamma function. We empirically analyze the impact of this adjustment in~\S\ref{sec:ablations}. 

We next outline a proposition, and prove in~\S\ref{app:propcom}, that demonstrates that reweighting using the adjusted proposal provably leads to better SNIS effective sample size (ESS).

\vspace{5pt}
\begin{mdframed}[style=MyFrame2]
\begin{restatable}{proposition}{propcom}
\label{prop:com}
Given an $\sethree$-invariant $\mu_{\text{target}}(x)$, consider the decomposition of a data point $x \in \R^{n \times 3}$ into its constituent mean-free component, $\bar{x} \in \R^{(n-1) \times 3} \hookrightarrow \R^{n \times 3}$ and center of mass $c \in \R^3$,  $x = \bar{x} + c$, where $c \sim \gN(0, \sigma^2)$. Now, assume both the proposal $p_{\theta}(x)$ and the adjusted proposal $p^c_{\theta}(x)$ factorize independently over the mean-free component and the center of mass. Then setting $p^c_{\theta}(x) = p_{\theta}(\bar{x}) \cdot 1/ \sigma \chi_3(\| c\|)$, leads to the following inequality on the effective sample size in the limit of $K \to \infty$:
\begin{equation}
     \text{ESS}\left(\frac{\mu_{\text{target}}(x)}{p_\theta(x)}\right) < \text{ESS}\left(\frac{\mu_{\text{target}}(x)}{p^c_{\theta}(x)}\right).
\end{equation}
\end{restatable}
\end{mdframed}

\subsection{Inference Time Scaling of Boltzmann Generators}
\label{sec:jarzynski}
\looseness=-1
Given a trained BG with proposal flow $p_{\theta}$, the self-normalized importance sampling estimator suffers from a large variance of importance weights as the dimensionality and complexity of $\mu_{\text{target}}(x)$ grows in large molecular systems. We aim to address this bottleneck by proposing an inference time scaling algorithm that anneals samples $x^i \sim p_{\theta}(x)$ --- and corresponding unnormalized importance weights $w(x^i)$ --- in a continuous manner towards $\mu_{\text{target}}$.

\xhdr{Improved sampling via annealing}
We leverage a class of methods that fall under non-equilibrium sampling to improve the base proposal flow samples. One of the simplest instantiations of this idea is to use annealed Langevin dynamics with reweighting through a continuous-time variant of Annealed Importance Sampling (AIS) \citep{neal2001annealed}. Concretely, we consider the following SDE that drives proposal samples towards the target Boltzmann density:
\begin{equation}
    \label{eqn:langevin_sde}
     d x_{\tau} = - \epsilon_{\tau} \nabla \E_{\tau}(x_{\tau }) d \tau  + \sqrt{2 \epsilon_{\tau}}dW_{\tau},
\end{equation}
where $\epsilon_{\tau} \geq 0$ is a time-dependent diffusion coefficient and $W_{\tau}$ is the standard Wiener process. We distinguish $\tau$, from $t$ used in the context of training $p_{\theta}$, as the time variable that evolves initial proposal samples at $\tau=0$ towards the target at $\tau=1$. The energy interpolation $\E_t$ is a design choice, and we opt for a simple linear interpolant $\E_t = (1-\tau) \E_0 + \tau \E_1$, and set $\gE_0(x) = -\log p_{\theta}(x)$. We highlight that unlike past work in pure sampling~\citep{mate2023learning,albergo_nets_2024} which use the prior energy $\E_0(x) = -\log p_0(x)$, our design affords the significantly more informative proposal given by the pre-trained normalizing flow $p_{\theta}$. As such, there is no need for \emph{additional learning}, with the annealing process extending the inference capabilities of the Boltzmann generator $\mu_{\theta}(x)$.

To resample or compute observables with the transported samples, we use the well-known and celebrated \emph{Jarzynski's equality}, that enables the calculation of equilibrium statistics from non-equilibrium processes. We recall the result, originally derived in \citet{jarzynski1997nonequilibirum}, and recently re-derived in continuous-time in the context of learned sampling algorithm by \citet{vargas2024transport,albergo_nets_2024}, that describes the importance weight time evolution.

\vspace{5pt}
\begin{mdframed}[style=MyFrame2]
\begin{proposition}[\citet{albergo_nets_2024}]
\label{prop:nets}
Let $(x_{\tau}, w_{\tau})$ solve the coupled system of SDE / ODE
\begin{align*}
    d x_{\tau} &= - \epsilon_{\tau} \nabla \E_{\tau}(x_{\tau }) d \tau  + \sqrt{2 \epsilon_{\tau}}dW_{\tau} \\
    d \log w_{\tau} &= - \partial_{\tau} \gE_{\tau}(x_{\tau}) d \tau \quad \text{with } x_0 \sim p_{\theta},  w_0 = 0
\end{align*}
then for any test function $\phi: \mathbb{R}^d \to \mathbb{R}$ we have
\begin{equation}
    \int_{\mathbb{R}^d} \phi(x) p_{\tau}(x) dx = \frac{\mathbb{E} [ w_{\tau} \phi(x_{\tau})]}{\mathbb{E}[w_{\tau}]}
\end{equation}
and
\begin{equation}
    \gZ_{\tau}/ \gZ_1 = \mathbb{E}[e^{w_{\tau}}] \quad (\text{Jarzynski's equality})
\end{equation}
\end{proposition}
\end{mdframed}
The final samples $x_{\tau=1}$ are then reweighted with the importance weights $w_{\tau=1}$, themselves lower variance than SNIS in conventional BGs. It is crucial to highlight that the prior is not directly constituent of this annealing process, but instead the learned proposal $p_{\theta}(x_0)$ acts as the initial distribution. It is precisely this learned proposal density that $d\log w_{\tau}$ evolves during the annealing process. Alnealed importance sampling can be considered a special case of sequential Monte Carlo (SMC) \citep{doucet2001sequential}. In SMC, resampling can occur at arbitrary times \(\tau\), typically using an ESS threshold as in adaptive resampling. Intuitively by resampling during the annealing process SMC can avoid particle redundancy in which all but a few particles have negligible weight. We state the full \nameshort sampling algorithm with adaptive resampling in \cref{alg:abgen}; to recover the \nameshort AIS variant we simply set \(\text{ESS}_\text{threshold} = -1.0\).

\vspace{10pt}
\begin{algorithm}[H]
\caption{\nameshort Sampling}
\label{alg:abgen}
\begin{algorithmic}[1]
\REQUIRE \# particles $K$, \# annealed distributions $N$, Energy annealing schedule $\mathcal{E}_\tau(x_{\tau})$
\STATE $x_0 \sim \mathcal{E}_0(x_0); \quad \Delta \gets 1 / N$
\FOR{$i = 1$ to $N$}
    \STATE $x_{\tau+\Delta} \gets x_\tau - \epsilon_\tau \nabla \mathcal{E}_\tau(x_\tau) d \tau + \sqrt{2 \epsilon_\tau} d W_\tau$
    \STATE $\log w_{\tau + \Delta} \gets \log w_\tau - \partial_\tau \mathcal{E}(x_\tau) d \tau$ \\
    \STATE $\tau \gets \tau + \Delta$
    \IF{$ \text{ESS} < \text{ESS}_\text{threshold}$}
        \STATE $x_\tau \gets \textsc{Resample}(x_\tau, w_\tau)$ 
        \STATE $w_\tau \gets 0$
    \ENDIF
\ENDFOR
\end{algorithmic}
\end{algorithm}

To simulate the Langevin SDE in \cref{eqn:langevin_sde}, and the corresponding importance weight evolution, we require the gradient of the energy interpolant:

\begin{equation*}
    \nabla \gE_{\tau}(x_{\tau}) = (1-\tau) \nabla(-\log p_{\theta}(x_{\tau})) + \tau \nabla \left(\frac{\E(x_{\tau})}{k_BT} \right),
\end{equation*}
which requires efficient gradient computation through the log-likelihood estimation under the normalizing flow $p_{\theta}$ as given by~\eqref{eqn:change_of_variables}. This presents the first point of distinction between finite flows and CNFs. The former class of flows trained using~\eqref{eqn:change_of_variables} gives fast exact likelihoods --- especially for our scalable non-equivariant TarFlow model. In contrast, CNFs must simulate~\eqref{eqn:cnf_and_log_prob_ode} and differentiate through an ODE solver to compute $\nabla \log p_{\theta} (x_{\tau})$ for each step of the Langevin SDE in~\eqref{eqn:langevin_sde}.
As a result, a TarFlow proposal is considerably cheaper to simulate and reweight with AIS than a CNF. In~\S\ref{app:alternate_paths} we present an alternate interpolant that does not require the proposal distribution during sampling which is appealing when only samples are needed but at the cost of more expensive computation of log weights. These paths are of interest in the setting of Boltzmann emulators and other generative models, and are of independent interest, but are not considered further in the context of \nameshort.

For improved numerical stability during annealing, and to further reduce computational footprint, we propose a strategy that eliminates the forward evolution of the initial proposal that already obtain high energy. Specifically, we can simulate a large number of samples via~\eqref{eqn:proposal_free_SDE} and threshold using an energy threshold $\gamma >0$, and evaluate the log weights of promising samples. We justify our strategy by first remarking a lower bound to the log partition function of $\mu_{\text{target}}$ using a Monte Carlo estimate,
\begin{align}
    \log \gZ &= \log \sE_{x \sim p_{\theta}(x)} \left[ \frac{\exp\left(\frac{-\E(x)}{k_BT}\right)}{p_{\theta}(x)} \right] \nonumber \\
    & \geq \sE_{x \sim p_{\theta}(x)} \left[\frac{-\E(x)}{k_BT} -\log p_{\theta}(x) \right] = \log \hat{\gZ}.
    \label{eqn:logZhat}
\end{align}
Plugging this estimate in the definition of the target Boltzmann distribution we get an upper bound,
\begin{align*}
   \log \mu_{\text{target}}(x) 
                            & \leq  \log \left(\frac{-\E(x)}{k_BT} \right) - \log \hat{\gZ} .
\end{align*}
An upper bound on $\mu_{\text{target}}(x)$ allows us to threshold samples using the energy function, $\gE(x) > \gamma$, of the target. Formally, this corresponds to truncating the target distribution $\hat{\mu}_{\text{target}}(x):= \sP\left(\mu_{\text{target}}(x) \geq \frac{\gamma}{\log \hat{\gZ}}\right)$ which places zero mass on high energy conformations. Correcting flow samples with respect to this truncated target introduces an additional bias into the self-normalized importance sampling estimate, which precisely corresponds to the difference in total variation distance between the two distributions $\text{TV} (\hat{\mu}_{\text{target}}, \mu_{\text{target}})$. We prove this result using an intermediate result in \cref{lemma:tvd_truncation} included in~\S\ref{app:proofs}. 

Our next theoretical result provides a prescriptive strategy of setting an appropriate threshold $\gamma$ as a function of the number of samples $K$ and effective sample size under $\hat{\mu}_{\text{target}}(x)$.

\vspace{5pt}
\begin{mdframed}[style=MyFrame2]
\begin{restatable}{proposition}{propenergythreshold}
\label{prop:energybias}
Given an energy threshold $\gE(x) > \gamma$, for $\gamma > 0$ large and the resulting truncated target distribution $\hat{\mu}_{\text{target}}(x):= \sP\left( \mu_{\text{target}}(x) \geq \frac{\gamma}{\log \hat{\gZ}} \right)$. Further, assume that the density of unnormalized importance weights w.r.t.\ to $\hat{\mu}_{\text{target}}$ is square integrable $(\hat{w}(x))^2 < \infty$. Given a tolerance $\rho = 1/ \text{ESS}$ and bias of the original importance sampling estimator in total variation $b= \text{TV}(\mu_{\theta}, \mu_{\text{target}})$, then the $\gamma$-truncation threshold with $K$-samples for $\text{TV}(\mu_{\theta}, \hat{\mu}_{\text{target}})$ is:
\begin{equation}
    \gamma \geq \frac{1}{\lambda} \log \left( \frac{K b}{12 \rho \sE[\exp(-\lambda X)]} \right) + \log \hat{\gZ}.
\end{equation}
\end{restatable}
\end{mdframed}
The proof for~\cref{prop:energybias} is located in~\S\ref{app:energybias}. \Cref{prop:energybias} allows us to appropriately set a energy threshold $\gamma$ as a function of tolerance $\rho$ that depends on ESS. In practice, this allows us to negotiate the amount of acceptable bias when dropping initial samples that obtain high-energy before any further AIS correction. Moreover, this gives a firmer theoretical foundation to existing practices of thresholding high importance weight samples~\citep{midgley2022flow,midgley2023se}.

Analogous to thresholding based on $\gE(x)$, we can also threshold by the probability under the proposal flow with truncation $\hat{p}_{\theta}(x) := \sP(p_{\theta}(x) \geq \delta)$, for small $\delta >0$. Essentially, this thresholding filters low probability samples under the model prior to any importance sampling. The additional bias incurred by performing such thresholding is theoretically analyzed in~\cref{prop:isbias} and presented in~\S\ref{app:isbias}. 

\section{Experiments}
\label{sec:experiments}

\begin{figure*}[t!]
    \centering
    \begin{subfigure}{0.19\linewidth}
        \centering
        \includegraphics[width=0.73\linewidth]{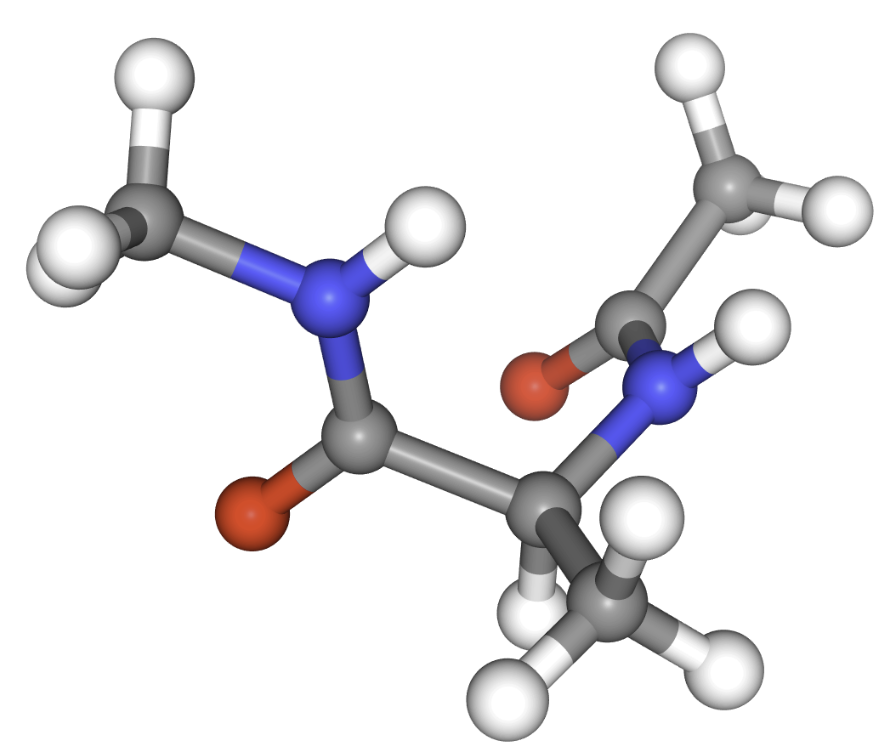}
        \caption{Alanine dipeptide}
        \label{fig:gen_aldp}
    \end{subfigure}
     \begin{subfigure}{0.19\linewidth}
        \centering
        \includegraphics[width=\linewidth]{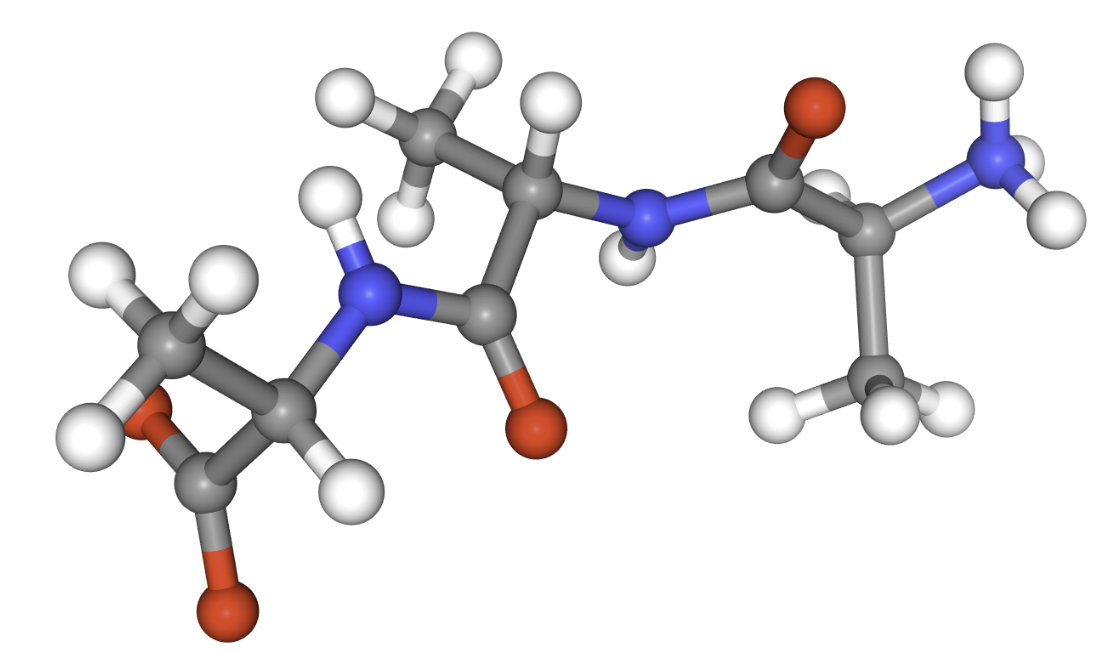}
        \caption{Trialanine}
        \label{fig:al3_gen}
    \end{subfigure}
     \begin{subfigure}{0.19\linewidth}
        \centering
        \includegraphics[width=\linewidth]{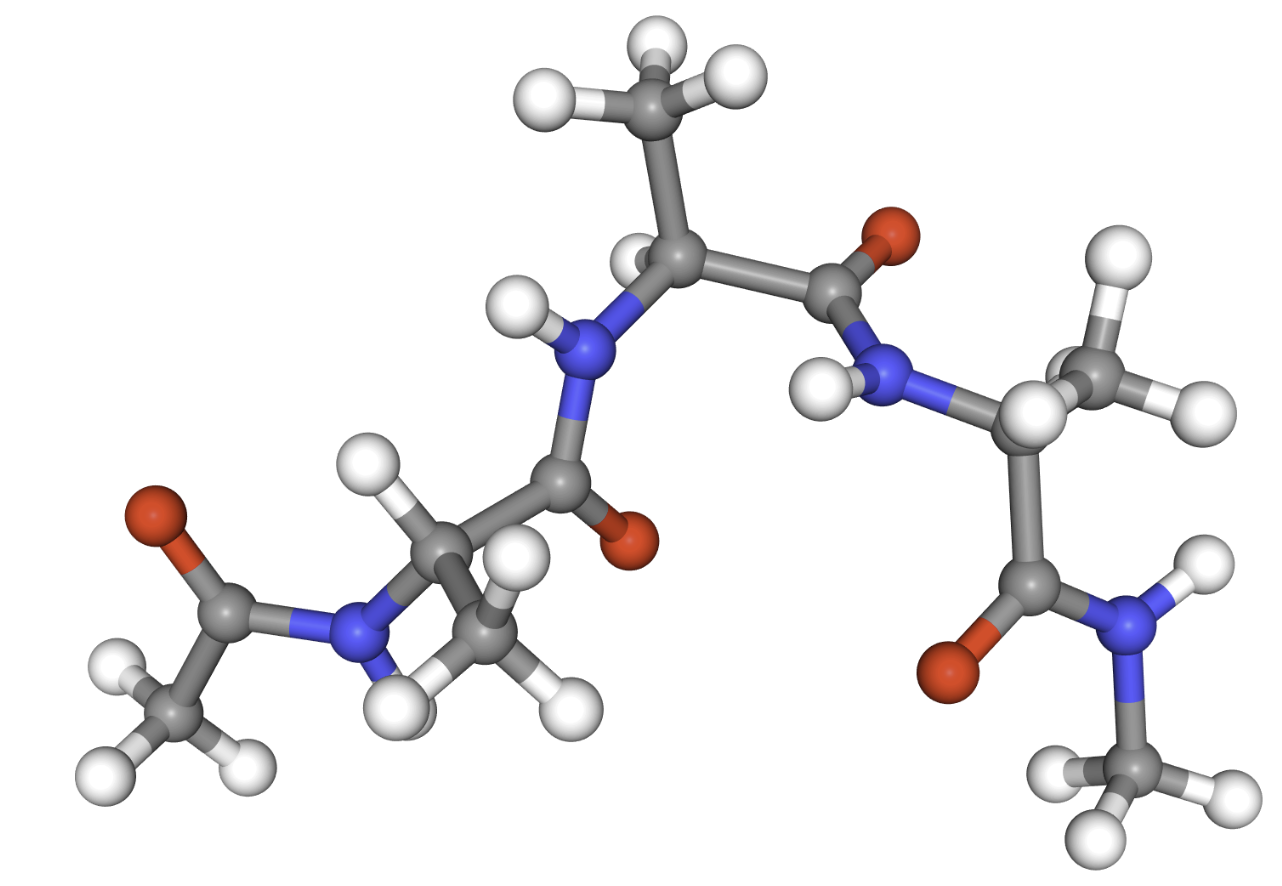}
        \caption{Alanine tetrapeptide}
        \label{fig:al4_gen}
    \end{subfigure}
    \begin{subfigure}{.19\linewidth}
    \centering
        \includegraphics[width=\linewidth]{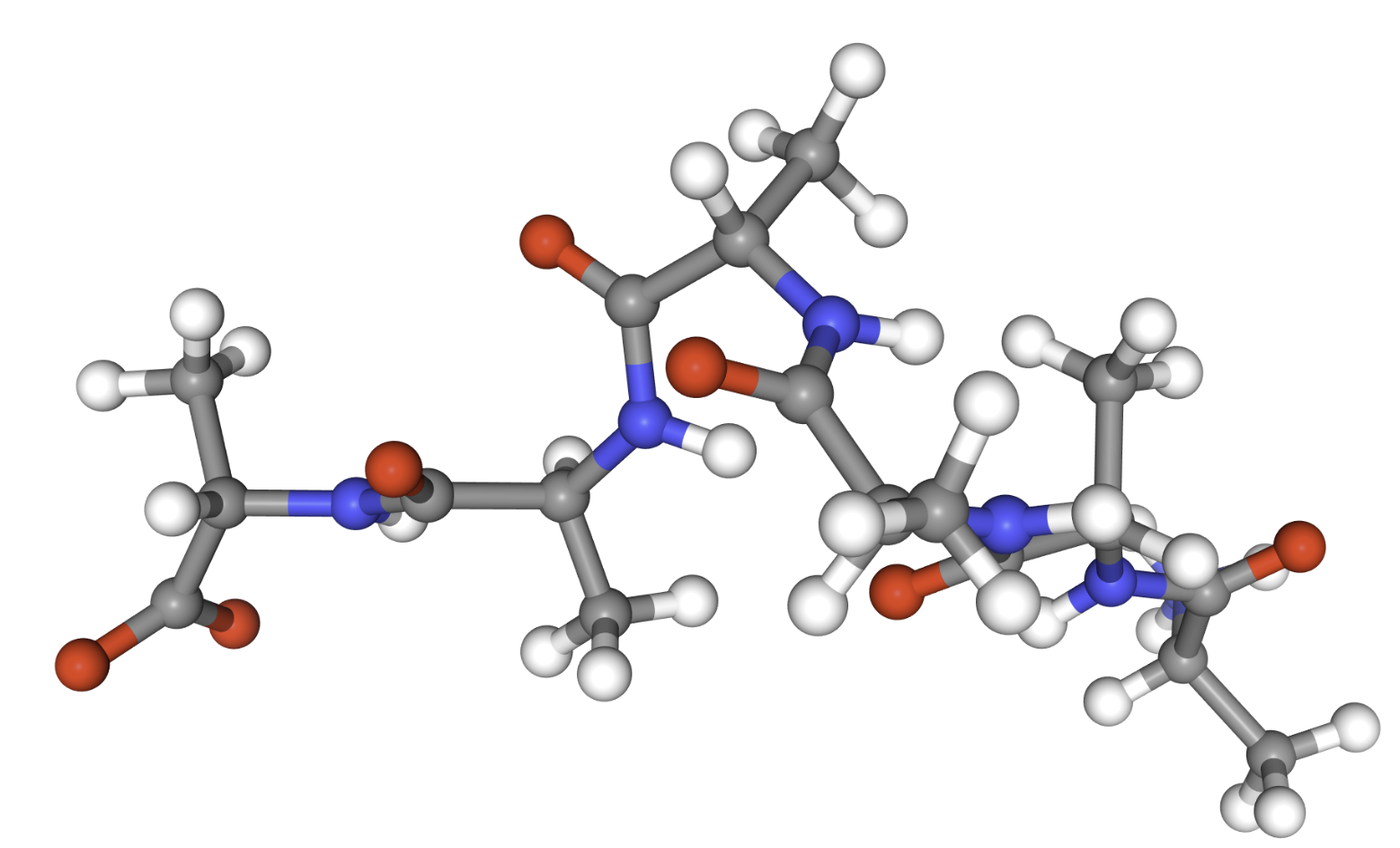}
        \caption{Hexa-alanine}
        \label{fig:al6_gen}
    \end{subfigure}
    \begin{subfigure}{.19\linewidth}
    \centering
        \includegraphics[width=0.8\linewidth]{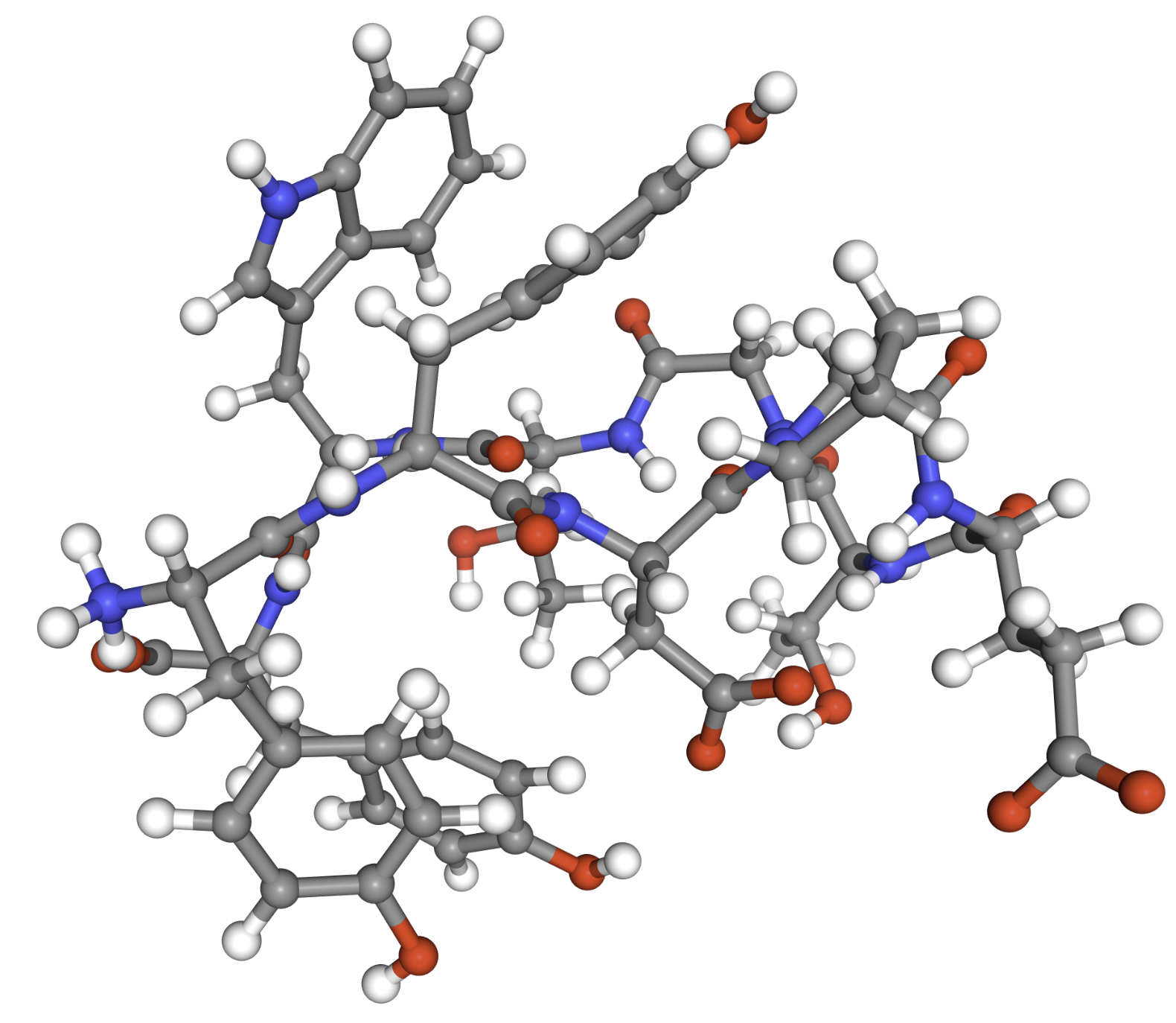}
        \caption{Chignolin}
        \label{fig:al10_gen}
    \end{subfigure}
    \vspace{-4pt}
    \caption{\small Samples generated by \nameshort on peptide systems ranging from 2 to 10 residues.}
    \label{fig:generated_samples}
    \vspace{-5pt}
\end{figure*}

\begin{table*}[ht!]
\centering
\caption{\small Quantitative results on alanine dipeptide and trialanine. Baseline methods presented with SNIS.}
\small
\label{tab:aldp_results}
\begin{tabular}{@{}lcccccc}
    \toprule
    & \multicolumn3c{Alanine dipeptide} &  \multicolumn3c{Trialanine} \\
    \cmidrule(lr){2-4}\cmidrule(lr){5-7}
    Algorithm $\downarrow$ & ESS $\uparrow$ & \emetric $\downarrow$ & \torusmetric $\downarrow$ & ESS $\uparrow$ & \emetric $\downarrow$ & \torusmetric $\downarrow$ \\
    \midrule
    $\sethree$-EACF & $<10^{-3}$ & 108.202 & 2.867 & --- & --- & --- \\
    ECNF & 0.119 & 0.419 & 0.311 & --- & --- & --- \\
    ECNF ++ (Ours) & \textbf{0.275 $\pm$ 0.010} &	0.914 $\pm$ 0.122&	\textbf{0.189 $\pm$ 0.019} & 0.003 $\pm$ 0.002&	2.206 $\pm$ 0.813& 0.962 $\pm$ 0.253 \\
    \midrule
    \nameshort AIS (Ours) & 0.030 $\pm$ 0.012 & 0.630 $\pm$ 0.249 & 0.418 $\pm$ 0.090 & \textbf{0.052 $\pm$ 0.013} & 0.797 $\pm$ 0.094 & \textbf{0.450 $\pm$ 0.043} \\
    \nameshort SMC (Ours) & --- & \textbf{0.412 $\pm$ 0.125} & 0.430 $\pm$ 0.100 & --- & \textbf{0.590 $\pm$ 0.267} & 0.455 $\pm$ 0.076 \\
    \bottomrule
    \end{tabular}
    \vspace{5pt}
\end{table*}

\begin{figure*}[ht!]
    \centering
    \begin{subfigure}{0.24\linewidth}
        \centering
        \includegraphics[width=\linewidth]{new_media/energies_eacf_al2.pdf}
        \caption{$\sethree$-EACF}
        \label{fig:ecnf_energy}
    \end{subfigure}
     \begin{subfigure}{0.24\linewidth}
        \centering
        \includegraphics[width=\linewidth]{new_media/energies_ecnf_al2.pdf}
        \caption{ECNF}
        \label{fig:ecnf_improved_energy}
    \end{subfigure}
     \begin{subfigure}{0.24\linewidth}
        \centering
        \includegraphics[width=\linewidth]{new_media/energies_ecnf++_al2.pdf}
        \caption{ECNF++}
        \label{fig:eacf_02_energy}
    \end{subfigure}
    \begin{subfigure}{.24\linewidth}
    \centering
        \includegraphics[width=\linewidth]{new_media/energies_tarflow_al2.pdf}
        \caption{\nameshort}
        \label{fig:al2_tarflow_energy}
    \end{subfigure}
    \vspace{-4pt}
    \caption{\small Energy histograms for baseline methods and \nameshort on alanine dipeptide dataset.}
    \vspace{-4pt}
    \label{fig:al2_energy_figs}
\end{figure*}

We evaluate \nameshort on small peptides using classical force-field energy functions, further experimental details are described in~\S\ref{app:exp_details}. \nameshort samples are generated by \cref{alg:abgen}, both with adaptive resampling (SMC) and without (AIS).

\xhdr{Datasets}
We consider small peptides composed of up to 6 alanine residues, with some systems additionally incorporating an acetyl group and an N-methyl group. All datasets are generated from a single MD simulation in implicit solvent using a classical force field. For each system, the first \SI{1}{\us} is used for training, the next \SI{0.2}{\us} for validation, and the remainder serves as the test set. Therefore, some metastable states may not be represented in the training set. An exception is alanine dipeptide, for which we use the dataset from~\citet{klein2024transferable}. In addition to the alanine systems, we also investigate the 138-atom peptide \emph{chignolin}, consisting of $10$ residues (\texttt{GYDPETGTWG}) and notable for it's formation of \(\beta\)-hairpin structure in water solvent \citet{honda200410}. We provide additional dataset details in~\S\ref{app:dataset_details}.

\xhdr{Baselines}
For baselines, we train prior state-of-the-art equivariant Boltzmann generators. Specifically, we train the exactly invertible and equivariant $\sethree$-augmented coupling flow~\citep{midgley2023se}, and the equivariant continuous normalizing flow (ECNF) employed in Transferable Boltzmann Generators~\citep{klein2024transferable}. We also include an improved variant, denoted ECNF++, as a stronger baselines; this uses a refined flow matching objective, larger network, and improved optimization hyperparameters, full details provided in ~\S\ref{app:ecnf_implementation_details} for full details. We note that both $\sethree$-EACH and ECNF to be equivariant to $\mathrm{E(3)}$ and hence generate samples of both global chiralities, which we resolve by applying a flip transformation as in \citet{klein2024transferable}, for further details and related results see \S\ref{app:additional_results}.

\xhdr{Metrics}
We report effective sample size (ESS) along with Wasserstein-2 distances on the energy distribution \emetric and dihedral angle torus \torusmetric. The energy distribution is highly sensitive to fine-grained details whereas the dihedral angles encode macrostructural information such as metastable state occupancy; full metric definitions are provided in \S\ref{app:exp_details}. Additional results for Wasserstein-2 distance on time-lagged independent component analysis (TICA) projections \ticametric are provided in \S\ref{app:additional_results}. We provide energy histograms in the main text whilst Ramachandran plots~\citep{ramachandran1963stereochemistry} detailing the mode coverage via dihedral angle distributions are presented in \S\ref{app:additional_results}.

\begin{figure}[H]
    \centering
     \begin{subfigure}{0.49\linewidth}
        \centering
        \includegraphics[width=\linewidth]{new_media/energies_ecnf++_al3.pdf}
        \label{fig:al3_ecnf_energy}
    \end{subfigure}
    \begin{subfigure}{.49\linewidth}
    \centering
         \includegraphics[width=\linewidth]{new_media/energies_tarflow_al3.pdf}
        \label{fig:al3_tarflow_energy}
    \end{subfigure}
    \vspace{-20pt}
    \caption{ \small Energy distribution histograms for baseline ECNF++ \textbf{(left)} and \nameshort \textbf{(right)} on trialanine dataset.}
    \label{fig:al3_energy_figs}
\end{figure}

\subsection{Results}
\label{sec:main_results}

We evaluate \nameshort and our baseline methods with quantitative metrics summarized in~\cref{tab:aldp_results} and~\cref{tab:main_results_wide}. Where $\pm$ is present three models are independently trained and sampled; unless otherwise stated \scione{4} particles are sampled. We provide examples of \nameshort generated samples in~\cref{fig:generated_samples}.

\xhdr{Alanine dipeptide} $\sethree$-EACF was originally trained on an alanine dipeptide dataset at \SI{800}{K}; we retrain on our more challenging \SI{300}{K} data using the original codebase of \citet{midgley2023se}. Despite the proposal distribution having good overlap with the MD data, we find the SNIS reweighted performance to be substantially  degraded at this lower temperature when using the same $0.2\%$ weight clipping threshold as the original work; see \S\ref{app:additional_results} for analysis of more aggressive clipping thresholds. The ESS and \torusmetric of ECNF++ outperform both \nameshort variants by a large margin, although the inverse is true for the \emetric. Furthermore, the original ECNF model trained by \citet{klein2024transferable} achieves superior \emetric to our proposed ECNF++ but inferior ESS and \torusmetric. These results are further substantiated by the energy distribution histograms in ~\cref{fig:al2_energy_figs}.

\begin{table*}[t!]
\centering
\caption{\small Quantitative results on alanine tetrapeptide and hexa-alanine. ECNF++ presented with SNIS.}
\small
\label{tab:main_results_wide}
\begin{tabular}{@{}lcccccc}
    \toprule
    Datasets $\rightarrow$ & \multicolumn3c{Alanine tetrapeptide} & \multicolumn3c{Hexa-alanine}  \\
    \cmidrule(lr){2-4}\cmidrule(lr){5-7}
    Algorithm $\downarrow$ & ESS $\uparrow$ & \emetric $\downarrow$ & \torusmetric $\downarrow$ & ESS $\uparrow$ & \emetric $\downarrow$ & \torusmetric $\downarrow$ \\
    \midrule
    ENCF++ (Ours) & 0.016 $\pm$ 0.001&	5.638 $\pm$ 0.483&	1.002 $\pm$ 0.061&		0.006 $\pm$ 0.001&	10.668 $\pm$ 0.285&	1.902 $\pm$ 0.055 \\
    \midrule 
    \nameshort AIS (Ours) & \textbf{0.046 $\pm$ 0.014} & \textbf{0.883 $\pm$ 0.213} & \textbf{0.866 $\pm$ 0.076} & \textbf{0.034 $\pm$ 0.015} & \textbf{1.021 $\pm$ 0.239} & \textbf{1.431 $\pm$ 0.085} \\
    \nameshort SMC (Ours) & --- & 1.027 $\pm$ 0.465 & 0.888 $\pm$ 0.114 & --- & 1.189 $\pm$ 0.357 & 1.444 $\pm$ 0.140 \\
    \bottomrule
    \end{tabular}
\end{table*}

\xhdr{Trialanine} Despite achieving acceptable performance on alanine dipeptide, ECNF was unable to scale to trialanine and was omitted. The computational cost of $\sethree$-EACF (c.f.~\cref{tab:training_times}) precluded it's consideration. The \nameshort variants are significantly stronger in all metrics compared to ECNF++, with \nameshort AIS achieving higher ESS and both AIS and SMC outperforming on \emetric and \torusmetric. However the performance of ECNF++ is acceptable, and constitutes the first CNF-based Boltzmann generator on a tripeptide system in Cartesian coordinates. There is no notable distinction in performance between \nameshort variants.

\begin{figure}[t]
    \captionsetup[subfigure]{aboveskip=0pt,belowskip=-1pt}
    \centering
     \begin{subfigure}{0.49\linewidth}
        \centering
        \includegraphics[width=\linewidth]{new_media/energies_ecnf++_al4.pdf}
        \label{fig:al4_ecnf_energy}
    \end{subfigure}
    \begin{subfigure}{.49\linewidth}
    \centering
         \includegraphics[width=\linewidth]{new_media/energies_tarflow_al4.pdf}
        \label{fig:al4_tarflow_energy}
    \end{subfigure}
    \vspace{-20pt}
    \caption{ \small Energy distribution histograms for baseline ECNF++ \textbf{(left)} and \nameshort \textbf{(right)} on alanine tetrapeptide dataset.}
    \label{fig:al4_energy_figs}
\end{figure}

\begin{figure}[t]
    \captionsetup[subfigure]{aboveskip=0pt,belowskip=-1pt}
    \centering
     \begin{subfigure}{0.49\linewidth}
        \centering
        \includegraphics[width=\linewidth]{new_media/energies_ecnf++_al6.pdf}
        \label{fig:al6_ecnf_energy}
    \end{subfigure}
    \begin{subfigure}{.49\linewidth}
    \centering
         \includegraphics[width=\linewidth]{new_media/energies_tarflow_al6.pdf}
        \label{fig:al6_tarflow_energy}
    \end{subfigure}
    \vspace{-20pt}
    \caption{ \small Energy distribution histograms for baseline ECNF++ \textbf{(left)} and \nameshort \textbf{(right)} on hexa-alanine dataset.}
    \label{fig:al6_energy_figs}
\end{figure}

\xhdr{Alanine tetrapeptide and hexa-alanine} At this scale the ECNF++ baseline diverges from the target distribution, reflected particularly in \emetric. In contrast, \nameshort is readily scalable up to hexapeptides, achieving greatly reduced \emetric on both datasets. As reweighted samples under \nameshort show extremely high overlap with the ground truth $\mu_{\text{target}}(x)$, we argue that \nameshort successfully solves these molecular systems in comparison to prior BGs. This conclusion is supported by the energy histograms in \cref{fig:al4_energy_figs} and \cref{fig:al6_energy_figs}, in which the SNIS reweighted SNIS does not approximate the MD data well, in contrast to the good alignment of \nameshort. Notably, the proposals for ECNF++ have good overlap with the target density, indicating the error to be introduced by the likelihood estimation itself.

\begin{figure}[!htb]
    \centering
     \begin{subfigure}{0.49\linewidth}
        \centering
        \includegraphics[width=\linewidth]{new_media/sampling_times.pdf}
    \end{subfigure}
    \begin{subfigure}{.49\linewidth}
    \centering
         \includegraphics[width=\linewidth]{new_media/torus_vs_timesteps.pdf}
    \end{subfigure}
    \vspace{-5pt}
    \caption{\small \textbf{Left:} GPU hours (NVIDIA L40S) for sampling and reweighting \scione{4} points. \textbf{Right:} \torusmetric on trialanine as a function of Langevin timestep discretization for both standard \(p_\theta(x)\) and center of mass adjusted proposal energy functions \(p^c_\theta(x)\).} 
    \label{fig:computational_cost_in_hours}
\end{figure}

\begin{figure}[!htb]
    \centering
     \begin{subfigure}{0.49\linewidth}
        \centering
        \includegraphics[width=\linewidth]{new_media/dists_tarflow_chignolin.pdf}
    \end{subfigure}
    \begin{subfigure}{.49\linewidth}
    \centering
        \raisebox{1.5pt}{
         \includegraphics[width=\linewidth]{new_media/energies_tarflow_chignolin.pdf}
         }
    \end{subfigure}
    \vspace{-5pt}
    \caption{ \small \nameshort interatomic distance histogram \textbf{(left)} and energy distribution histogram \textbf{(right)} for decapeptide chignolin (\texttt{GYDPETGTWG}) . SNIS 
    \(\mathcal{E}\text{‑}\mathcal{W}_2 = 12.046\), SMC  \(\mathcal{E}\text{‑}\mathcal{W}_2 = 3.571\).}
    \label{fig:chignolin_tarflow}
\end{figure}

\xhdr{Inference scaling}
To illustrate the scalability of \nameshort in relation to other methods we plot in~\cref{fig:computational_cost_in_hours} the GPU hours required by each method to sample \scione{4} points. We observe exponential scaling of inference time for ECNF++ as the size of the system grows, whilst \nameshort is less sensitive to system size and over an order of magnitude faster on the hexapeptide system. We additionally plot the \torusmetric on trialanine as a function of Langevin timestep granularity for \nameshort SMC both with and without the center of mass proposal energy adjustment, as stated \cref{eqn:com_adjust}. When the center of mass adjusted energy is employed we observe a strong inverse relationship between time discretization steps and \torusmetric, however without this adjustment there is no clear relationship. This evidences both the efficacy of the center of mass adjustment at improving reweighting as well as the potential of \nameshort for inference-time scaling --- a capability not present in the standard Boltzmann generator.

\subsection{Scaling to Decapeptide}
\label{sec:decapeptides}
We now apply \nameshort to the decapeptide chignolin. As no other method can scale to this system we report energy histograms and distance plots for \nameshort SMC only in~\cref{fig:chignolin_tarflow}. We observe success of \nameshort at matching the interatomic distance distribution. We additionally observe a strong overlap of SMC sample energy distribution despite the notably poor proposal overlap, providing further demonstration of the viability of the \nameshort approach to molecular system sampling. Our application of \nameshort to chignolin represents a significant step forwards in the scalability of BGs, where prior methods struggled on even alanine tetrapeptide, as observable in results for ECNF++ \emetric presented in \cref{tab:main_results_wide}.

\section{Related Work}
\label{sec:related_work}
Boltzmann generators (BGs)~\citep{noe2019boltzmann} have been applied to both free energy estimation~\citep{wirnsberger2020targeted, rizzi2023multimap, schebek2024efficient} and molecular sampling. Initially, BGs relied on system-specific representations, such as internal coordinates, to achieve relevant sampling efficiencies \citep{noe2019boltzmann, kohler2021smooth, midgley2022flow, kohler2023rigid, dibak2021temperature}. However, these representations are generally not transferable across different systems, leading to the development of BGs operating in Cartesian coordinates~\citep{klein2023equivariant, midgley2023se,klein2024transferable}. While this improves transferability, they are currently limited in scalability, struggling to extend beyond dipeptides. Scaling to larger systems typically requires sacrificing exact sampling from the target distribution  \citep{jing2022torsional,abdin2023pepflow,jing2024alphafold,lewis2024scalable}.
An alternative to direct sampling from $\mu_{\text{target}}(x)$ is to generate samples iteratively by learning large steps in time~\citep{schreiner2023implicit, fu2023simulate, klein2023timewarp, diez2024boltzmann, jing2024generative, daigavane2024jamun} to accelerate methods such as molecular dynamics via coarse-graining.

\xhdr{Amortized sampling}
The field of sampling has seen renewed interest with the rise of generative models. In particular, the use of diffusion-based samplers has seen rapid application with a plethora of approaches exploiting the favorable theoretical properties of mode-mixing of diffusion models~\citep{berner2022optimal,vargas2023denoising,richter2023improved,zhang2021path,vargas2024transport}.
While initial approaches focused on simulation-based dynamics, including both overdamped and underdamped Langevin~\citep{blessing2025underdamped,chen_sequential_2024}, it is expected that simulation-free methods that also exploit diffusion properties~\citep{akhound2024iterated,huang2021schrodinger,de2024target} are an attractive opportunity to tackle larger-scale systems due to their scalability. 
Finally, flow-based models have also been employed for sampling with classical flows augmenting MCMC~\citep{arbel2021annealed, gabrie2021efficient, matthews2022continual,midgley2022flow,hagemann2023generalized}, and through CNFs that construct ODE bridges, such as linear interpolants between the prior and target~\citep{mate2023learning}, and more general bridges that rely on satisfying the mass transport equations~\citep{tian2024liouville,fan2024path}.

\section{Conclusion}
\label{sec:conclusion}
In this paper, we introduce \nameshort an extension to the Boltzmann generator framework that scales inference through the use of annealing processes. Unlike past BGs, in \nameshort, we scale training using a non-equivariant transformer-based TarFlow architecture with soft equivariance penalties to $6$ peptides. In terms of limitations, using non-equilibrium sampling as presented in \nameshort does not enjoy easy application to CNFs due to expensive simulation, which limits the use of modern flow matching methods in a \nameshort context. Considering hybrid approaches that mix CNFs through distillation to an invertible architecture or consistency-based objectives is thus a natural direction for future work. Finally, considering other classes of scalable generative models such as autoregressive ones which also permit exact likelihoods is also a ripe direction for future work.

\section*{Acknowledgements}
\label{sec:acknowledgements}
The authors thank Damien Ferbach, Tara Akhound-Sadegh, Lars Holdijk, Kacper Kapuśniak, Kirill Neklyudov, Michael Albergo, and Majdi Hassan for insightful conversations and feedback. In addition, the authors thank Paul Skaluba for constructive comments on Proposition 1 of an older draft.

The authors acknowledge funding from UNIQUE, CIFAR, NSERC, Intel, and Samsung. The research was enabled in part by computational resources provided by the Digital Research Alliance of Canada (\url{https://alliancecan.ca}), Mila (\url{https://mila.quebec}), and NVIDIA.
AJB is partially supported by an NSERC Post-doc
fellowship. This research is partially supported by the EPSRC Turing AI World-Leading Research Fellowship No. EP/X040062/1 and EPSRC AI Hub No. EP/Y028872/1.

\section*{Impact Statement}
\label{sec:impact_statement}

This work studies sampling from Boltzmann densities, a problem of general interest in machine learning and AI4Science that arises both in pure statistical modeling and within applications. We highlight the training Boltzmann generators on molecular tasks are in turn applicable to drug and material discovery. While we do not foresee immediate negative impacts of our advances in this area, we encourage due caution whilst scaling to prevent their potential misuse.

\clearpage

\bibliography{clean}

@article{honda200410,
  title={10 residue folded peptide designed by segment statistics},
  author={Honda, Shinya and Yamasaki, Kazuhiko and Sawada, Yoshito and Morii, Hisayuki},
  journal={Structure},
  volume={12},
  number={8},
  pages={1507--1518},
  year={2004},
  publisher={Elsevier}
}

@article{abdin2023pepflow,
	title        = {PepFlow: direct conformational sampling from peptide energy landscapes through hypernetwork-conditioned diffusion},
	author       = {Abdin, Osama and Kim, Philip M},
	year         = {2023},
	journal      = {bioRxiv},
	pages        = {2023--06}
}

@article{abramson2024accurate,
	title        = {Accurate structure prediction of biomolecular interactions with AlphaFold 3},
	author       = {Abramson, Josh and Adler, Jonas and Dunger, Jack and Evans, Richard and Green, Tim and Pritzel, Alexander and Ronneberger, Olaf and Willmore, Lindsay and Ballard, Andrew J and Bambrick, Joshua and others},
	year         = {2024},
	journal      = {Nature},
	pages        = {1--3}
}

@article{agapiou2017importance,
	title        = {Importance sampling: Intrinsic dimension and computational cost},
	author       = {Agapiou, Sergios and Papaspiliopoulos, Omiros and Sanz-Alonso, Daniel and Stuart, Andrew M},
	year         = {2017},
	journal      = {Statistical Science},
	pages        = {405--431}
}

@inproceedings{akhound2024iterated,
	title        = {Iterated Denoising Energy Matching for Sampling from Boltzmann Densities},
	author       = {Akhound-Sadegh, Tara and Rector-Brooks, Jarrid and Bose, Joey and Mittal, Sarthak and Lemos, Pablo and Liu, Cheng-Hao and Sendera, Marcin and Ravanbakhsh, Siamak and Gidel, Gauthier and Bengio, Yoshua and Malkin, Nikolay and Tong, Alexander},
	booktitle    = {International Conference on Machine Learning (ICML)},
    year         = {2024},
}

@article{albergo_building_2023,
	title        = {Building Normalizing Flows with Stochastic Interpolants},
	author       = {Albergo, Michael S. and {Vanden-Eijnden}, Eric},
	year         = {2023},
	journal      = {International Conference on Learning Representations (ICLR)}
}

@inproceedings{albergo_nets_2024,
	title        = {NETS: A Non-Equilibrium Transport Sampler},
	author       = {Michael S. Albergo and Eric Vanden-Eijnden},
	year         = {2025},
	booktitle    = {International Conference on Machine Learning (ICML)}
}

@inproceedings{alexey2020image,
	title        = {An image is worth 16x16 words: Transformers for image recognition at scale},
	author       = {Alexey, Dosovitskiy},
	year         = {2021},
	booktitle    = {International Conference on Learning Representations (ICLR)}
}

@inproceedings{arbel2021annealed,
	title        = {Annealed flow transport monte carlo},
	author       = {Arbel, Michael and Matthews, Alex and Doucet, Arnaud},
	year         = {2021},
	booktitle    = {International Conference on Machine Learning},
	pages        = {318--330},
	organization = {PMLR}
}

@article{berner2022optimal,
	title        = {An optimal control perspective on diffusion-based generative modeling},
	author       = {Berner, Julius and Richter, Lorenz and Ullrich, Karen},
	year         = {2024},
	journal      = {Transactions on Machine Learning Research (TMLR)}
}

@inproceedings{blessing2025underdamped,
	title        = {Underdamped diffusion bridges with applications to sampling},
	author       = {Blessing, Denis and Berner, Julius and Richter, Lorenz and Neumann, Gerhard},
	year         = {2025},
	booktitle    = {International Conference on Learning Representations (ICLR)}
}

@article{bose2021equivariant,
	title        = {Equivariant finite normalizing flows},
	author       = {Bose, Avishek Joey and Brubaker, Marcus and Kobyzev, Ivan},
	year         = {2021},
	journal      = {arXiv}
}

@article{buch2011complete,
	title        = {Complete reconstruction of an enzyme-inhibitor binding process by molecular dynamics simulations},
	author       = {Buch, Ignasi and Giorgino, Toni and De Fabritiis, Gianni},
	year         = {2011},
	journal      = {Proceedings of the National Academy of Sciences},
	volume       = {108},
	number       = {25},
	pages        = {10184--10189}
}

@article{chen_neural_2018,
	title        = {Neural Ordinary Differential Equations},
	author       = {Chen, Ricky T. Q. and Rubanova, Yulia and Bettencourt, Jesse and Duvenaud, David K},
	year         = {2018},
	journal      = {Neural Information Processing Systems (NIPS)}
}

@inproceedings{chen_sequential_2024,
	title        = {Sequential Controlled Langevin Diffusions},
	author       = {Junhua Chen and Lorenz Richter and Julius Berner and Denis Blessing and Gerhard Neumann and Anima Anandkumar},
	year         = {2025},
	booktitle    = {International Conference on Learning Representations (ICLR)}
}

@article{daigavane2024jamun,
	title        = {JAMUN: Transferable Molecular Conformational Ensemble Generation with Walk-Jump Sampling},
	author       = {Daigavane, Ameya and Vani, Bodhi P and Saremi, Saeed and Kleinhenz, Joseph and Rackers, Joshua},
	year         = {2024},
	journal      = {arXiv}
}

@article{de2024target,
	title        = {Target score matching},
	author       = {De Bortoli, Valentin and Hutchinson, Michael and Wirnsberger, Peter and Doucet, Arnaud},
	year         = {2024},
	journal      = {arXiv}
}

@article{Deringer_2020,
	title        = {Modelling and understanding battery materials with machine-learning-driven atomistic simulations},
	author       = {Deringer, Volker L},
	year         = {2020},
	month        = {oct},
	journal      = {Journal of Physics: Energy},
	volume       = {2},
	number       = {4},
	pages        = {041003},
	doi          = {10.1088/2515-7655/abb011}
}

@article{dibak2021temperature,
	title        = {Temperature steerable flows and {Boltzmann} generators},
	author       = {Dibak, Manuel and Klein, Leon and Kr\"amer, Andreas and No\'e, Frank},
	year         = {2022},
	month        = {Oct},
	journal      = {Phys. Rev. Res.},
	volume       = {4},
	pages        = {L042005},
	doi          = {10.1103/PhysRevResearch.4.L042005},
	issue        = {4},
	numpages     = {6}
}

@article{diez2024boltzmann,
	title        = {Boltzmann priors for Implicit Transfer Operators},
	author       = {Diez, Juan Viguera and Schreiner, Mathias and Engkvist, Ola and Olsson, Simon},
	year         = {2025},
	journal      = {International Conference on Learning Representations (ICLR)}
}

@article{dinh2016density,
	title        = {Density estimation using {Real NVP}},
	author       = {Dinh, Laurent and Sohl-Dickstein, Jascha and Bengio, Samy},
	year         = {2017},
	journal      = {International Conference on Learning Representations (ICLR)}
}

@inproceedings{domingo2024adjoint,
	title        = {Adjoint Matching: Fine-tuning Flow and Diffusion Generative Models with Memoryless Stochastic Optimal Control},
	author       = {Carles Domingo-Enrich and Michal Drozdzal and Brian Karrer and Ricky T. Q. Chen},
	year         = {2025},
	booktitle    = {International Conference on Representation Learning (ICLR)}
}

@book{doucet2001sequential,
	title        = {Sequential Monte Carlo methods in practice},
	author       = {Doucet, Arnaud and De Freitas, Nando and Gordon, Neil James and others},
	year         = {2001},
	volume       = {1},
	number       = {2}
}

@article{eastman2017openmm,
	title        = {OpenMM 7: Rapid development of high performance algorithms for molecular dynamics},
	author       = {Eastman, Peter and Swails, Jason and Chodera, John D and McGibbon, Robert T and Zhao, Yutong and Beauchamp, Kyle A and Wang, Lee-Ping and Simmonett, Andrew C and Harrigan, Matthew P and Stern, Chaya D and others},
	year         = {2017},
	journal      = {PLoS computational biology},
	volume       = {13},
	number       = {7},
	pages        = {e1005659}
}

@inproceedings{esser2024scalingrectifiedflowtransformers,
	title        = {Scaling Rectified Flow Transformers for High-Resolution Image Synthesis},
	author       = {Patrick Esser and Sumith Kulal and Andreas Blattmann and Rahim Entezari and Jonas Müller and Harry Saini and Yam Levi and Dominik Lorenz and Axel Sauer and Frederic Boesel and Dustin Podell and Tim Dockhorn and Zion English and Kyle Lacey and Alex Goodwin and Yannik Marek and Robin Rombach},
	year         = {2024},
	booktitle    = {International Conference on Machine Learning (ICML)}
}

@inproceedings{fan2024path,
	title        = {Path-guided particle-based sampling},
	author       = {Fan, Mingzhou and Zhou, Ruida and Tian, Chao and Qian, Xiaoning},
	year         = {2024},
	booktitle    = {International Conference on Machine Learning (ICML)}
}

@article{flamary2021pot,
	title        = {POT: Python Optimal Transport},
	author       = {R{\'e}mi Flamary and Nicolas Courty and Alexandre Gramfort and Mokhtar Z. Alaya and Aur{\'e}lie Boisbunon and Stanislas Chambon and Laetitia Chapel and Adrien Corenflos and Kilian Fatras and Nemo Fournier and L{\'e}o Gautheron and Nathalie T.H. Gayraud and Hicham Janati and Alain Rakotomamonjy and Ievgen Redko and Antoine Rolet and Antony Schutz and Vivien Seguy and Danica J. Sutherland and Romain Tavenard and Alexander Tong and Titouan Vayer},
	year         = {2021},
	journal      = {Journal of Machine Learning Research},
	volume       = {22},
	number       = {78},
	pages        = {1--8}
}

@article{fu2023simulate,
	title        = {Simulate time-integrated coarse-grained molecular dynamics with multi-scale graph networks},
	author       = {Fu, Xiang and Xie, Tian and Rebello, Nathan J and Olsen, Bradley and Jaakkola, Tommi S},
	year         = {2023},
	journal      = {Transactions on Machine Learning Research}
}

@article{gabrie2021efficient,
	title        = {Efficient {Bayesian} sampling using normalizing flows to assist {Markov} chain {Monte Carlo} methods},
	author       = {Gabri{\'e}, Marylou and Rotskoff, Grant M and Vanden-Eijnden, Eric},
	year         = {2021},
	journal      = {arXiv}
}

@article{garcia2021n,
	title        = {{E(n)} equivariant normalizing flows},
	author       = {Garcia Satorras, Victor and Hoogeboom, Emiel and Fuchs, Fabian and Posner, Ingmar and Welling, Max},
	year         = {2021},
	journal      = {Neural Information Processing Systems (NeurIPS)}
}

@inproceedings{grathwohl_ffjord_2018,
	title        = {{FFJORD:} Free-form Continuous Dynamics for Scalable Reversible Generative Models},
	author       = {Will Grathwohl and Ricky T. Q. Chen and Jesse Bettencourt and Ilya Sutskever and David Duvenaud},
	year         = {2019},
	booktitle    = {International Conference on Representation Learning (ICLR)}
}

@book{hagemann2023generalized,
	title        = {Generalized normalizing flows via Markov chains},
	author       = {Hagemann, Paul Lyonel and Hertrich, Johannes and Steidl, Gabriele},
	year         = {2023}
}

@article{huang2021schrodinger,
	title        = {Schr{\"o}dinger-{F}{\"o}llmer sampler: sampling without ergodicity},
	author       = {Huang, Jian and Jiao, Yuling and Kang, Lican and Liao, Xu and Liu, Jin and Liu, Yanyan},
	year         = {2021},
	journal      = {arXiv}
}

@article{Hutchinson01011990,
	title        = {A stochastic estimator of the trace of the influence matrix for laplacian smoothing splines},
	author       = {M.F. Hutchinson},
	year         = {1990},
	journal      = {Communications in Statistics - Simulation and Computation},
	volume       = {19},
	number       = {2},
	pages        = {433--450},
	doi          = {10.1080/03610919008812866},
	eprint       = {https://doi.org/10.1080/03610919008812866}
}

@article{jarzynski1997nonequilibirum,
	title        = {Nonequilibrium Equality for Free Energy Differences},
	author       = {Jarzynski, C.},
	year         = {1997},
	month        = {Apr},
	journal      = {Phys. Rev. Lett.},
	volume       = {78},
	pages        = {2690--2693},
	doi          = {10.1103/PhysRevLett.78.2690},
	issue        = {14},
	numpages     = {0}
}

@article{jing2022torsional,
	title        = {Torsional diffusion for molecular conformer generation},
	author       = {Jing, Bowen and Corso, Gabriele and Chang, Jeffrey and Barzilay, Regina and Jaakkola, Tommi},
	year         = {2022},
	journal      = {Advances in Neural Information Processing Systems},
	volume       = {35},
	pages        = {24240--24253}
}

@inproceedings{jing2024alphafold,
	title        = {AlphaFold meets flow matching for generating protein ensembles},
	author       = {Jing, Bowen and Berger, Bonnie and Jaakkola, Tommi},
	year         = {2024},
	booktitle    = {International Conference on Machine Learning (ICML)}
}

@inproceedings{jing2024generative,
	title        = {Generative modeling of molecular dynamics trajectories},
	author       = {Jing, Bowen and St{\"a}rk, Hannes and Jaakkola, Tommi and Berger, Bonnie},
	year         = {2024},
	booktitle    = {Neural Information Processing Systems (NeurIPS)}
}

@inproceedings{karczewski2024diffusion,
	title        = {Diffusion Models as Cartoonists! The Curious Case of High Density Regions},
	author       = {Karczewski, Rafa{\l} and Heinonen, Markus and Garg, Vikas},
	year         = {2024},
	booktitle    = {International Conference on Learning Representations (ICLR)}
}

@article{klein2023equivariant,
	title        = {Equivariant flow matching},
	author       = {Leon Klein and Andreas Krämer and Frank Noé},
	year         = {2023},
	journal      = {Neural Information Processing Systems (NeurIPS)}
}

@article{klein2023timewarp,
	title        = {Timewarp: Transferable acceleration of molecular dynamics by learning time-coarsened dynamics},
	author       = {Klein, Leon and Foong, Andrew YK and Fjelde, Tor Erlend and Mlodozeniec, Bruno and Brockschmidt, Marc and Nowozin, Sebastian and No{\'e}, Frank and Tomioka, Ryota},
	year         = {2023},
	journal      = {Neural Information Processing Systems (NeurIPS)}
}

@inproceedings{klein2024transferable,
	title        = {Transferable Boltzmann Generators},
	author       = {Klein, Leon and No{\'e}, Frank},
	year         = {2024},
	booktitle    = {Advances in Neural Information Processing Systems}
}

@article{kohler2020equivariant,
	title        = {Equivariant flows: exact likelihood generative learning for symmetric densities},
	author       = {K{\"o}hler, Jonas and Klein, Leon and No{\'e}, Frank},
	year         = {2020},
	journal      = {International Conference on Machine Learning (ICML)}
}

@inproceedings{kohler2021smooth,
	title        = {Smooth Normalizing Flows},
	author       = {K\"{o}hler, Jonas and Kr\"{a}mer, Andreas and Noé, Frank},
	year         = {2021},
	booktitle    = {Advances in Neural Information Processing Systems},
	volume       = {34},
	pages        = {2796--2809},
	editor       = {M. Ranzato and A. Beygelzimer and Y. Dauphin and P.S. Liang and J. Wortman Vaughan}
}

@article{kohler2023rigid,
	title        = {Rigid body flows for sampling molecular crystal structures},
	author       = {K{\"o}hler, Jonas and Invernizzi, Michele and De Haan, Pim and No{\'e}, Frank},
	year         = {2023},
	journal      = {International Conference on Machine Learning (ICML)}
}

@article{lee2021universal,
	title        = {Universal approximation using well-conditioned normalizing flows},
	author       = {Lee, Holden and Pabbaraju, Chirag and Sevekari, Anish Prasad and Risteski, Andrej},
	year         = {2021},
	journal      = {Advances in Neural Information Processing Systems},
	volume       = {34},
	pages        = {12700--12711}
}

@article{lewis2024scalable,
	title        = {Scalable emulation of protein equilibrium ensembles with generative deep learning},
	author       = {Lewis, Sarah and Hempel, Tim and Jim{\'e}nez Luna, Jos{\'e} and Gastegger, Michael and Xie, Yu and Foong, Andrew YK and Garc{\'\i}a Satorras, Victor and Abdin, Osama and Veeling, Bastiaan S and Zaporozhets, Iryna and others},
	year         = {2024},
	journal      = {bioRxiv},
	pages        = {2024--12}
}

@article{lindorff2011fast,
	title        = {How fast-folding proteins fold},
	author       = {Lindorff-Larsen, Kresten and Piana, Stefano and Dror, Ron O and Shaw, David E},
	year         = {2011},
	journal      = {Science},
	volume       = {334},
	number       = {6055},
	pages        = {517--520}
}

@article{lipman_flow_2022,
	title        = {Flow Matching for Generative Modeling},
	author       = {Lipman, Yaron and Chen, Ricky T. Q. and {Ben-Hamu}, Heli and Nickel, Maximilian and Le, Matt},
	year         = {2023},
	journal      = {International Conference on Learning Representations (ICLR)}
}

@article{liu_rectified_2022,
	title        = {Rectified Flow: A Marginal Preserving Approach to Optimal Transport},
	author       = {Liu, Qiang},
	year         = {2022},
	journal      = {arXiv}
}

@misc{liu2024instaflowstephighqualitydiffusionbased,
	title        = {InstaFlow: One Step is Enough for High-Quality Diffusion-Based Text-to-Image Generation},
	author       = {Xingchao Liu and Xiwen Zhang and Jianzhu Ma and Jian Peng and Qiang Liu},
	year         = {2024},
	eprint       = {2309.06380},
	archiveprefix = {arXiv},
	primaryclass = {cs.LG}
}

@inproceedings{loshchilov2017decoupled,
	title        = {Decoupled weight decay regularization},
	author       = {Loshchilov, I and Hutter, F},
	year         = {2017},
	booktitle    = {International Conference on Learning Representations (ICLR)}
}

@article{mate2023learning,
	title        = {Learning Interpolations between Boltzmann Densities},
	author       = {B{\'a}lint M{\'a}t{\'e} and Fran{\c{c}}ois Fleuret},
	year         = {2023},
	journal      = {Transactions on Machine Learning Research},
	issn         = {2835-8856},
	note         = {}
}

@article{matsumoto2002molecular,
	title        = {Molecular dynamics simulation of the ice nucleation and growth process leading to water freezing},
	author       = {Matsumoto, Masakazu and Saito, Shinji and Ohmine, Iwao},
	year         = {2002},
	journal      = {Nature},
	volume       = {416},
	number       = {6879},
	pages        = {409--413}
}

@article{matthews2022continual,
	title        = {Continual repeated annealed flow transport Monte Carlo},
	author       = {Matthews, Alex and Arbel, Michael and Rezende, Danilo Jimenez and Doucet, Arnaud},
	year         = {2022},
	journal      = {International Conference on Machine Learning (ICML)}
}

@book{mcbook,
	title        = {Monte Carlo theory, methods and examples},
	author       = {Art B. Owen},
	year         = {2013}
}

@article{midgley2022flow,
	title        = {Flow annealed importance sampling bootstrap},
	author       = {Midgley, Laurence Illing and Stimper, Vincent and Simm, Gregor NC and Sch{\"o}lkopf, Bernhard and Hern{\'a}ndez-Lobato, Jos{\'e} Miguel},
	year         = {2023},
	journal      = {International Conference on Learning Representations (ICLR)}
}

@article{midgley2023se,
	title        = {{SE(3)} Equivariant Augmented Coupling Flows},
	author       = {Midgley, Laurence I and Stimper, Vincent and Antor{\'a}n, Javier and Mathieu, Emile and Sch{\"o}lkopf, Bernhard and Hern{\'a}ndez-Lobato, Jos{\'e} Miguel},
	year         = {2023},
	journal      = {Neural Information Processing Systems (NeurIPS)}
}

@article{neal2001annealed,
	title        = {Annealed importance sampling},
	author       = {Neal, Radford M},
	year         = {2001},
	journal      = {Statistics and computing},
	volume       = {11},
	pages        = {125--139}
}

@article{noe2009constructing,
	title        = {Constructing the equilibrium ensemble of folding pathways from short off-equilibrium simulations},
	author       = {No{\'e}, Frank and Sch{\"u}tte, Christof and Vanden-Eijnden, Eric and Reich, Lothar and Weikl, Thomas R},
	year         = {2009},
	journal      = {Proceedings of the National Academy of Sciences},
	volume       = {106},
	number       = {45},
	pages        = {19011--19016}
}

@article{noe2019boltzmann,
	title        = {Boltzmann generators: Sampling equilibrium states of many-body systems with deep learning},
	author       = {No{\'e}, Frank and Olsson, Simon and K{\"o}hler, Jonas and Wu, Hao},
	year         = {2019},
	journal      = {Science},
	volume       = {365},
	number       = {6457},
	pages        = {eaaw1147}
}

@article{papamakarios2017masked,
	title        = {Masked autoregressive flow for density estimation},
	author       = {Papamakarios, George and Pavlakou, Theo and Murray, Iain},
	year         = {2017},
	journal      = {Advances in neural information processing systems},
	volume       = {30}
}

@article{parrinello1980crystal,
	title        = {Crystal structure and pair potentials: A molecular-dynamics study},
	author       = {Parrinello, Michele and Rahman, Aneesur},
	year         = {1980},
	journal      = {Physical review letters},
	volume       = {45},
	number       = {14},
	pages        = {1196}
}

@article{ramachandran1963stereochemistry,
	title        = {Stereochemistry of polypeptide chain configurations},
	author       = {Ramachandran, G N and Ramakrishnan, C and Sasisekharan, V},
	year         = {1963},
	journal      = {Journal of Molecular Biology},
	pages        = {95--99}
}

@article{rezende2015variational,
	title        = {Variational inference with normalizing flows},
	author       = {Rezende, Danilo and Mohamed, Shakir},
	year         = {2015},
	journal      = {International Conference on Machine Learning (ICML)}
}

@article{richter2023improved,
	title        = {Improved sampling via learned diffusions},
	author       = {Richter, Lorenz and Berner, Julius and Liu, Guan-Horng},
	year         = {2024},
	journal      = {International Conference on Learning Representations (ICLR)}
}

@article{rizzi2023multimap,
	title        = {Free energies at QM accuracy from force fields via multimap targeted estimation},
	author       = {Rizzi, Andrea and Carloni, Paolo and Parrinello, Michele},
	year         = {2023},
	journal      = {Proceedings of the National Academy of Sciences}
}

@article{satorras2021n,
	title        = {E (n) equivariant graph neural networks},
	author       = {Satorras, V{\i}ctor Garcia and Hoogeboom, Emiel and Welling, Max},
	year         = {2021},
	journal      = {International Conference on Machine Learning (ICML)}
}

@article{schebek2024efficient,
	title        = {Efficient mapping of phase diagrams with conditional Boltzmann Generators},
	author       = {Schebek, Maximilian and Invernizzi, Michele and No{\'e}, Frank and Rogal, Jutta},
	year         = {2024},
	journal      = {Machine Learning: Science and Technology}
}

@inproceedings{schreiner2023implicit,
	title        = {Implicit Transfer Operator Learning: Multiple Time-Resolution Models for Molecular Dynamics},
	author       = {Mathias Schreiner and Ole Winther and Simon Olsson},
	year         = {2023},
	booktitle    = {Thirty-seventh Conference on Neural Information Processing Systems}
}

@inproceedings{skreta2024superposition,
	title        = {The Superposition of Diffusion Models Using the Itô Density Estimator},
	author       = {Skreta, Marta and Atanackovic, Lazar and Bose, Avishek Joey and Tong, Alexander and Neklyudov, Kirill},
	year         = {2025},
	booktitle    = {International Conference on Learning Representations (ICLR)}
}

@article{teshima2020coupling,
	title        = {Coupling-based invertible neural networks are universal diffeomorphism approximators},
	author       = {Teshima, Takeshi and Ishikawa, Isao and Tojo, Koichi and Oono, Kenta and Ikeda, Masahiro and Sugiyama, Masashi},
	year         = {2020},
	journal      = {Advances in Neural Information Processing Systems},
	volume       = {33},
	pages        = {3362--3373}
}

@inproceedings{tian2024liouville,
	title        = {Liouville flow importance sampler},
	author       = {Tian, Yifeng and Panda, Nishant and Lin, Yen Ting},
	year         = {2024},
	booktitle    = {International Conference on Machine Learning (ICML)}
}

@article{tong_conditional_2023,
	title        = {Improving and Generalizing Flow-Based Generative Models with Minibatch Optimal Transport},
	author       = {Tong, Alexander and Fatras, Kilian and Malkin, Nikolay and Huguet, Guillaume and Zhang, Yanlei and {Rector-Brooks}, Jarrid  and Wolf, Guy and Bengio, Yoshua},
	year         = {2024},
	journal      = {Transactions on Machine Learning Research}
}

@article{vargas2023denoising,
	title        = {Denoising diffusion samplers},
	author       = {Vargas, Francisco and Grathwohl, Will and Doucet, Arnaud},
	year         = {2023},
	journal      = {International Conference on Learning Representations (ICLR)}
}

@article{vargas2024transport,
	title        = {Transport meets Variational Inference: Controlled {Monte Carlo} Diffusions},
	author       = {Francisco Vargas and Shreyas Padhy and Denis Blessing and Nikolas N{\"u}sken},
	year         = {2024},
	journal      = {International Conference on Learning Representations (ICLR)}
}

@article{wirnsberger2020targeted,
	title        = {Targeted free energy estimation via learned mappings},
	author       = {Wirnsberger, Peter and Ballard, Andrew J and Papamakarios, George and Abercrombie, Stuart and Racani{\`e}re, S{\'e}bastien and Pritzel, Alexander and Jimenez Rezende, Danilo and Blundell, Charles},
    journal      = {J. Chem. Phys.},
	year         = {2020},
}

@article{zhai2024normalizing,
	title        = {Normalizing flows are capable generative models},
	author       = {Zhai, Shuangfei and Zhang, Ruixiang and Nakkiran, Preetum and Berthelot, David and Gu, Jiatao and Zheng, Huangjie and Chen, Tianrong and Bautista, Miguel Angel and Jaitly, Navdeep and Susskind, Josh},
	year         = {2024},
	journal      = {arXiv}
}

@article{zhang2021path,
	title        = {Path integral sampler: a stochastic control approach for sampling},
	author       = {Zhang, Qinsheng and Chen, Yongxin},
	year         = {2022},
	journal      = {International Conference on Learning Representations (ICLR)}
}
\bibliographystyle{style/icml2025}

\appendix
\onecolumn

\section{Alternate Paths}
\label{app:alternate_paths}

\subsection{Proposal Free Langevin Dynamics}
\looseness=-1
We can also modify the Langevin SDE in~\eqref{eqn:langevin_sde} to include an additional drift term $\nu_{\tau}(x_{\tau}) \in \R^d$ as follows:
\begin{equation*}
    d x_{\tau} = - \epsilon_{\tau} \nabla \gE_t(x_{\tau }) d \tau + \nu_{\tau} ( x_{\tau}) d \tau + \sqrt{2 \epsilon_{\tau}}dW_{\tau}.
\end{equation*}
\looseness=-1
Under perfect drift $\nu_{\tau}(\tau)$ the log weights do not change and there is no need for correction. For imperfect drift the corresponding coupled ODE time-evolution of log-weights $d \log w_{\tau}$ needed to apply AIS was derived in NETS~\citep[Proposition 3]{albergo_nets_2024}:
\begin{equation*}
     d w_{\tau} = \nabla \cdot \nu_{\tau} ( x_{\tau}) d\tau - \nabla \E_{\tau}(x_{\tau}) \cdot \nu_{\tau}( x_{\tau}) d\tau - \partial_{\tau} \gE_{\tau}(x_{\tau}) d\tau.
\end{equation*}
In contrast to learning a drift as done in NETS~\citep{albergo_nets_2024} we now illustrate that a judicious choice of $\nu_{\tau}(x_{\tau})$ eliminates the need to compute the gradient of log-likelihood under the proposal. For instance, we can choose $\nu_{\tau}(x_{\tau}) = \epsilon_{\tau} \nabla \E_{\tau}(x_{\tau}) - \epsilon_{\tau} \nabla \left( \frac{\E (x_{\tau})}{k_BT}\right)$, which by straightforward calculation gives the following SDE:
\begin{align}
     d x_{\tau} & = - \epsilon_{\tau} \nabla \gE_t(x_{\tau }) d \tau + \nu_{\tau} ( x_{\tau}) d \tau + \sqrt{2 \epsilon_{\tau}}dW_{\tau} \nonumber \\
     &=  - \epsilon_{\tau} \nabla \left( \frac{\E (x_{\tau})}{k_BT}\right) d\tau  + \sqrt{2 \epsilon_{\tau}}dW_{\tau}.
     \label{eqn:proposal_free_SDE}
\end{align}
\looseness=-1
This new SDE greatly simplifies the simulation of samples $x_{\tau}$ as it is independent of the proposal energy $\nabla \gE_0(x_{\tau}) = -\nabla \log p_{\theta}(x_{\tau})$. However, the log weights ODE still requires the computation of the gradient of the proposal energy. The form of~\eqref{eqn:proposal_free_SDE} suggests the possibility of massively parallel simulation schemes under a regular normalizing flow and a CNF. However, due to simulatio the log weights remains expensive for CNFs due to the need to compute the divergence operator in~\eqref{eqn:cnf_and_log_prob_ode}. Furthermore, while recent advances in divergence-free density estimation via the It$\hat{\text{o}}$ density estimator~\citep{skreta2024superposition,karczewski2024diffusion} might appear attractive we show that the log density under this estimator is necessarily biased and may limit the fidelity of self-normalized importance sampling incurs non-negotiable added bias. For ease of presentation, we present this theoretical investigation in~\S\ref{app:ito_filtering} and characterize the added bias in~\cref{prop:weights_bias}. In totality, this limits the application of continuous BG's to only the conventional IS setting, unlike finite flows like TarFlow which can benefit from non-equilibrium transport and AIS.

\section{Proofs}
\label{app:proofs}

\subsection{Proof of~\cref{prop:com}}
\label{app:propcom}

\vspace{5pt}
\begin{mdframed}[style=MyFrame2]
\propcom*
\end{mdframed}

\begin{proof}
\looseness=-1
Recall the definition of effective sample size using Kish's formula, both $p_{\theta}(x)$ and the adjusted proposal $p^c_{\theta}(x)$:
\begin{align*}
    \text{ESS}\left( \frac{\mu_{\text{target}(x)}}{p_{\theta}(x)} \right) &= \frac{1}{\sum^K_i (\bar{w}(x^i))^2} = \frac{\left( \sum^K_i w(x^i)\right)^2}{\sum_i^K w(x^i)^2} = \frac{\left( \sum^K_i \mu_{\text{target}}(x^i) / p_{\theta}(x^i) \right)^2}{\sum_i^K w(x^i)^2} \\
     \text{ESS}\left( \frac{\mu_{\text{target}(x)}}{p^c_{\theta}(x)} \right) &= \frac{1}{\sum^K_i (\bar{w}^c(x^i))^2} = \frac{\left( \sum^K_i w^c(x^i)\right)^2}{\sum_i^K w^c(x^i)^2} = \frac{\left( \sum^K_i \mu_{\text{target}}(x^i) / p^c_{\theta}(x^i) \right)^2}{\sum_i^K  \left(\mu_{\text{target}}(x^i) / p^c_{\theta}(x^i) \right)^2}.
\end{align*}

We can rewrite the weights for the regular proposal's ESS calculation as follows,
\begin{align*}
    w(x^i) &=  \frac{\mu_{\text{target}}(\bar{x}^i + c )}{p_{\theta}(\bar{x}^i + c)} = \frac{\mu_{\text{target}}(\bar{x}^i)}{p_{\theta}(\bar{x}^i)p(c)} \\
    w(x^i) & = w(\bar{x}^i) \cdot w(c)
\end{align*}
where we exploited the translation invariance of $\mu_{\text{target}}$ to remove the center of mass $c$ and also the independence between the mean $\bar{x}$ and $c$ in the proposal. Since $c \sim \gN(0, \sigma^2)$ we can write $p(c) = p(\| c\| , \theta, \phi)$ in spherical coordinates to follow a scaled Chi distribution $\| c\| \sim \sigma \chi_3 (\|c \|)$ with angular components that follow independent uniform distributions $(\theta, \phi) \sim \gU(\theta) \gU(\phi)$. Fixing canonical angular components $(\theta, \phi)$ we have $p(c) = \sigma \chi_3 (\| c\|)$ and $w(c) = 1/\sigma \chi_3 (\| c\|)$.

For the adjusted proposal we set $p(c) = 1/ \sigma \chi_3$ which gives the following weights:
\begin{equation}
     w^c(x^i) =  \frac{\mu_{\text{target}}(\bar{x}^i + c )}{p^c_{\theta}(\bar{x}^i + c)} = \frac{\mu_{\text{target}}(\bar{x}^i)}{p_{\theta}(\bar{x}^i) p(c)} = \frac{\mu_{\text{target}}(\bar{x}^i)}{p_{\theta}(\bar{x}^i)} = w(\bar{x}^i).
\end{equation}

We now seek to prove that:
\begin{equation}
    \frac{\left(\sum_{i=1}^{K} w(\bar{x}^i) w(c)\right)^2}
{\sum_{i=1}^{K} w(\bar{x}^i)^2 w(c)^2} < \frac{\left(\sum_{i=1}^{K} w(\bar{x}^i) \right)^2}
{\sum_{i=1}^{K} w(\bar{x}^i)^2}.
\label{eq:bound_statement}
\end{equation}

Now, denote $X^i = w(\bar{x}^i)$ and $Y^i = w(c)$ random variables over the sample index $i$. By construction, \(X^i\) and \(Y^i\) are independent for each \(i\). Furthermore, all $(X^i, Y^i)$ pairs are i.i.d. for $i \in [K]$.

We may now formalize equation \ref{eq:bound_statement} as proving

\begin{equation}
  \frac{\bigl(\mathbb{E}[X\,Y]\bigr)^2}{\mathbb{E}[(X\,Y)^2]}
  \;<\;
  \frac{\bigl(\mathbb{E}[X]\bigr)^2}{\mathbb{E}[X^2]}
  .
  \label{eq:bound_population}
\end{equation}

Because \(X\) and \(Y\) are independent for each sample we know:
\begin{equation}
   \mathbb{E}[X\,Y] 
   \;=\;
   \mathbb{E}[X]\;\mathbb{E}[Y],
   \quad\quad
   \mathbb{E}[(X\,Y)^2]
   \;=\;\mathbb{E}[X^2]\;\mathbb{E}[Y^2].
\end{equation}

Hence
\begin{equation}
  \frac{\bigl(\mathbb{E}[X\,Y]\bigr)^2}{\mathbb{E}[(X\,Y)^2]}
  \;=\;
  \frac{\bigl(\mathbb{E}[X]\bigr)^2}{\mathbb{E}[X^2]}
  \;\times\;
  \frac{\bigl(\mathbb{E}[Y]\bigr)^2}{\mathbb{E}[Y^2]}.
\end{equation}

If \(\operatorname{Var}(Y)>0\), then \(\mathbb{E}[Y^2] > (\mathbb{E}[Y])^2\).  Consequently,
  \begin{equation}
    0 \;<\;\frac{\bigl(\mathbb{E}[Y]\bigr)^2}{\mathbb{E}[Y^2]}
    \;<\;1.
  \end{equation}
  
Thus, at the level of population expectations,  
\begin{equation}
  \frac{\bigl(\mathbb{E}[X\,Y]\bigr)^2}{\mathbb{E}[(X\,Y)^2]}
  \;=\;
  \frac{\bigl(\mathbb{E}[X]\bigr)^2}{\mathbb{E}[X^2]}
  \;\times\;
  \underbrace{\frac{\bigl(\mathbb{E}[Y]\bigr)^2}{\mathbb{E}[Y^2]}}_{<1}
  \;<\;
  \frac{\bigl(\mathbb{E}[X]\bigr)^2}{\mathbb{E}[X^2]}
  \;\;.
\end{equation}
Therefore, applying the adjustment is strictly better by ESS than unadjusted.

\end{proof}

\subsection{Proof of~\cref{lemma:tvd_truncation}}
\label{app:lemma_truncation}

We first prove a useful lemma that computes the total variation distance between the original distribution of the normalizing flow $p_{\theta}$ and the truncated distribution $\hat{p}_{\theta}$ before proving the propositions.

\begin{mdframed}[style=MyFrame2]
\begin{lemma}
Let $p_{\theta}$ be a generative and denote $\hat{p}_{\theta}(x)$ the $\delta$-truncated distribution such that $\hat{p}_{\theta}(x): = \sP( p_{\theta}(x) \geq \delta)$, for a small $\delta > 0$. Define the constant $\beta  = \sP(p_{\theta}(x) < \delta)$ as the event where the truncation occurs. Then the total variation distance between the generative model and its truncated distribution is $ \text{TV}(p_{\theta}, \hat{p}_{\theta}) = \beta$.
\label{lemma:tvd_truncation}
\end{lemma}
\end{mdframed}

\begin{proof}
We begin by first characterizing the total variation distance between flow after correction with importance sampling $p(x)$ with truncated distribution $\hat{p}(x)$. Recall that the truncated distribution is defined as follows:

\begin{equation}
    \hat{p}(x): = \sP( p(x) \geq \delta)  = \frac{p(x) \mathbb{I}\{ p(x) \geq \delta\}}{\int  \mathbb{I}\{ p(x) \geq \delta\} p(x) dx},
\end{equation}
where $\mathbb{I}$ is the indicator function. Denote the events $\alpha = \sP(X \geq \delta)$ and $\beta = \sP(X < \delta)$ for the random variance $X \sim p(x)$. Clearly, $\alpha + \beta =1$ and $\alpha =\int  \mathbb{I}\{ \mu(x) \geq \delta\} p(x) dx $. Now consider the total variation distance between these two distributions:

\begin{equation}
    \text{TV}(p, \hat{p}) =  \sup_{\phi \in \Phi } \left| \sE_{x \sim p(x)}[\phi(x) ]- \sE_{\hat{x} \sim \hat{p}(x)} [\phi(\hat{x})]  \right| = \frac{1}{2} \int | p(x) - \hat{p}(x)| dx.
\end{equation}
\looseness=-1
where $\Phi = \{ \phi: \| \phi\|_{\infty} \leq 1\}$.
Next we break up the event space into two regions $R_1$ and $R_2$ which correspond to the events $p(x) < \delta$ and $p(x) \geq \delta$ respectively. Now consider the total variation distance in the region $R_1$ whereby construction $\hat{p}(x) = 0$, 
\begin{equation}
 \frac{1}{2} \int_{R_1} | p(x) - \hat{p}(x)| dx = \frac{1}{2}\int_{R_1} p(x) dx = \frac{\beta}{2}.
\end{equation}
A similar computation on $R_2$ gives,
\begin{equation}
 \frac{1}{2} \int_{R_2} | p_{\theta}(x) - \hat{p}_{\theta}(x)| dx =   \frac{1}{2} \int_{R_2} \left| p(x) - \frac{p(x)}{\alpha} \right| dx = \frac{1}{2}\int_{R_2} p(x) \left | 
 1 - \frac{1}{\alpha}\right| dx = \frac{\alpha (\frac{1}{\alpha} -1)}{2} = \frac{\beta}{2},
\end{equation}
where we exploited the fact that $\hat{p_{\theta}}(x) = \frac{p_{\theta}(x)}{\alpha}$ in the first equality and that $\alpha = \int_{R_2} p_{\theta}(x) dx$ in the second equality. Combining these results we get the full total variation distance:
\begin{equation}
     \text{TV}(p, \hat{p}) =  \frac{1}{2} \int | p(x) - \hat{p}(x)| dx = \frac{1}{2} \int_{R_1} | p(x) - \hat{p}(x)| dx + \frac{1}{2} \int_{R_2} | p(x) - \hat{p}(x)| dx = \beta.
\end{equation}
Thus the $\text{TV}(p, \hat{p}) = \beta$ and $0$ in the trivial case where $\alpha = 1$ and the truncated distribution are the same.
\end{proof}

\subsection{Proof of~\cref{prop:energybias}}
\label{app:energybias}

\vspace{5pt}
\begin{mdframed}[style=MyFrame2]
\propenergythreshold*
\end{mdframed}

\begin{proof}
We start by recalling a well-known result stating the bias of self-normalized importance sampling found in~\citet[Theorem 2.1]{agapiou2017importance} using $K$ samples from the proposal $\mu(x)$:
\begin{equation}
    \sup_{\| \phi \|_{\infty} \leq 1} \left| \sE \left[ \mu_{\theta}^K (\phi) -\mu_{\text{target}} (\phi)  \right] \right| \leq \frac{12 \rho}{K}, \quad \rho \approx \frac{K}{\text{ESS}} = \frac{K\sum^K_jw(x^j)^2 }{\left(\sum_i^K w(x^i)\right)^2} 
\end{equation} 
where the terms $\mu_{\theta}^K(\phi) = \sum_i^K \bar{w}(x^i) \phi(x^i)$ is the self-normalized importance estimator of $\mu_{\text{target}}$ with samples drawn according to $x^i \sim p_{\theta}(x)$ and $\|\phi(x)\| \leq 1 $ is a bounded test function.

\looseness=-1
By truncating using an energy threshold $\E(x) < \gamma$, for a large $\gamma > 0$, we truncate the support of $\mu_{\text{target}}(x)$ by cutting off low probability regions that constitute high-energy configurations. More precisely, we have $\hat{\mu}_{\text{target}}:= \sP\left(\mu_{\text{target}}(x) \geq \frac{\gamma}{\log \hat{\gZ}}\right)$, where $\log \hat{\gZ}$ is as defined in~\eqref{eqn:logZhat}. Note that $\hat{\mu}_{\text{target}}(x)$ is absolutely continuous w.r.t. to $\mu_{\text{target}}$ as the support is contained up to modulo measure zero sets. The importance sampling error incurred by using $\hat{\mu}_{\text{target}}$ can be bounded as follows:
\begin{align}
    \sup_{\| \phi \|_{\infty} \leq 1} \left| \sE \left[ \mu_{\theta}^K (\phi) -\mu_{\text{target}} (\phi)  \right] \right| & \leq   \sup_{\| \phi \|_{\infty} \leq 1} \left| \sE \left[ \mu_{\theta}^K (\phi) -\hat{\mu}_{\text{target}} (\phi)  \right]  \right| +  \sup_{\| \phi \|_{\infty} \leq 1} \left| \sE \left[ \hat{\mu}_{\text{target}}(\phi) - \mu_{\text{target}}(\phi)\right] \right| \\
    & \leq \frac{12 \hat{\rho}}{K} + \beta_1 \\
    & \leq \frac{12 \rho}{K} + \beta_1.
\end{align}
\looseness=-1
The first inequality follows from the triangle inequality. Here we note that $\hat{\rho}$ is the ESS which corresponds to using importance weights computed with respect to the truncated target $\hat{\mu}_{\text{target}}$ rather than $\mu_{\text{target}}$. The constant $\beta_1 = \text{TV}(\hat{\mu}_{\text{target}}, \mu_{\text{target}})$ and follows from an application of \cref{lemma:tvd_truncation}. Further, note that $\rho \geq \hat{\rho}$ since ESS must increase---and thereby $\hat{\rho}$ decreases---as the distributional overlap between the two distributions decreases.
Now observe, $\beta_1 = \sP\left(X < \frac{\gamma}{\log \hat{\gZ}}\right)$, where samples follow the law $X \sim \hat{\mu}_{\text{target}}(x)$. Then a 
 direct application of Chernoff's inequality gives us $\sP\left(X < \frac{\gamma}{\log \hat{\gZ}}\right) = \beta_1 \leq \exp\left(\frac{\lambda \gamma}{\log \hat{\gZ}}\right) \sE[\exp\left(-\lambda X\right)]$. Thus the additional bias incurred is, 
\begin{equation}
    \sup_{\| \phi \|_{\infty} \leq 1} \left| \sE \left[ \hat{\mu}_{\theta}^K (\phi) -\mu_{\text{target}} (\phi)  \right] \right| \leq \frac{12 \rho}{K} + \beta_1 \leq \frac{12 \rho}{K} +  \exp\left(\frac{\lambda \gamma}{\log \hat{\gZ}}\right) \sE[\exp(-\lambda X)].
\end{equation}
Where the term $\sE[\exp(-\lambda X)]$ is the moment generating function. Setting $b:= \text{TV}(\mu^K_{\theta}, \mu_{\text{target}})$, then we have
\begin{equation}
   \gamma \geq \frac{1}{\lambda} \log \left( \frac{K b}{12 \rho \sE[\exp(-\lambda X)]} \right) + \log \hat{\gZ}.
\end{equation}
\end{proof}

\subsection{Proof of \cref{prop:isbias}}
\label{app:isbias}

\vspace{5pt}
\begin{mdframed}[style=MyFrame2]
\begin{restatable}{proposition}{propbias}
\label{prop:isbias}
\looseness=-1
Assume that the density of the model $p_{\theta}$ after importance sampling $\mu_{\theta}$ is absolutely continuous with respect to the target $\mu_{\text{target}}$. Further, assume that the density of unnormalized importance weights is square integrable $(w(x))^2 < \infty$. Given a tolerance $\rho = 1/ \text{ESS}$ of the original importance sampling estimator under $\mu_{\theta}$ and bias of the importance sampling estimator in total variation $b= \text{TV}(\mu_{\theta}, \mu_{\text{target}})$, then the $\delta$-truncation for the truncated distribution $\hat{p}_{\theta}(x) := \sP(p_{\theta}(x) \geq \delta)$ threshold with $K$-samples is:
\begin{equation}
    \delta \geq \frac{1}{\lambda} \log \left( \frac{K b}{12 \rho \sE[\exp(-\lambda X)]} \right).
\end{equation}
\end{restatable}
\end{mdframed}
\begin{proof}
\looseness=-1
We aim to bound the total variation distance $\text{TV}( \hat{\mu}_{\theta}^K, \mu_{\text{target}})$ of using the truncated distribution $\sP(p_{\theta}(x) > \delta)$ by again recalling the bias of self-normalized importance sampling using $K$ samples from $\mu_{\theta}(x)$:
\begin{equation}
    \sup_{\| \phi \|_{\infty} \leq 1} \left| \sE \left[ \mu_{\theta}^K (\phi) -\mu_{\text{target}} (\phi)  \right] \right| \leq \frac{12 \rho}{K}, \quad \rho \approx \frac{K}{\text{ESS}} = \frac{K\sum^K_jw(x^j)^2 }{\left(\sum_i^K w(x^i)\right)^2} 
\end{equation} 

where the terms $\mu_{\theta}^K(\phi) = \sum_i^K \bar{w}(x^i) \phi(x^i)$ is the self-normalized importance estimator of $\mu_{\text{target}}$ with samples drawn according to $x^i \sim p_{\theta}(x)$ and $\|\phi(x)\| \leq 1 $ is a bounded test function. We next characterize the error introduced by using the truncated distribution $\hat{p}_{\theta}$ for importance sampling in place of $p_{\theta}$ by first defining the truncated $K$-sample self-normalized importance estimator $\hat{\mu}_{\theta}^K (\phi)= \sum^K_j \bar{w}(x^j)\phi(x^j)$, where $x^j \sim \hat{p}_{\theta}(x)$. Specifically, we bound the total variation distance:
\begin{align}
    \text{TV}(\mu_{\theta}, \hat{\mu}_{\theta}) &=  \sup_{\| \phi \|_{\infty} \leq 1} \left| \sE \left[ \mu_{\theta}^K (\phi) - \hat{\mu}_{\theta}^K (\phi)  \right] \right| \\
    &=   \sup_{\| \phi \|_{\infty} \leq 1} \left| \sE_{x^i \sim p_{\theta}} \left[ \sum^K_{i=1}\bar{w}(x^i) \phi(x^i) \right]  - \sE_{x^j \sim \hat{p}_{\theta}} \left[\sum^K_{j=1} \bar{w}(x^j)\phi(x^j)  \right] \right| \\
   & = \frac{1}{2}\left( \sE_{x^i \sim p_{\theta}} \left[\sum^K_{i=1} \bar{w}(x^i) \right]  - \sE_{x^j \sim \hat{p}_{\theta}} \left[ \sum^K_{j=1} \bar{w}(x^j)  \right] \right)
\end{align}
Here in the second equality, we used the fact that the test function is bounded $|| \phi \||_{\infty} \leq 1$ 
Next, we apply~\cref{lemma:tvd_truncation} and leverage the fact that the self-normalized weights are also bounded and achieve a bound on the total variation distance,
\begin{align}
    \text{TV}(\mu, \hat{\mu}) & =  \frac{1}{2}\left( \sE_{x^i \sim p_{\theta}} \left[ \sum^K_{i=1} \bar{w}(x^i) \right]  - \sE_{x^j \sim \hat{p}_{\theta}} \left[\sum^K_{j=1} \bar{w}(x^j)  \right] \right) \\
    & = \beta_2,
\end{align}

where $\beta_2$ is the probability mass $\sP(X< \delta)$ when $X \sim p_{\theta}(x)$. Like previously, the overall error can be bounded using the triangle inequality
\begin{align}
    \sup_{\| \phi \|_{\infty} \leq 1} \left| \sE \left[ \mu_{\theta}^K (\phi) -\mu_{\text{target}} (\phi)  \right] \right| & \leq   \sup_{\| \phi \|_{\infty} \leq 1} \left| \sE \left[ \hat{\mu}^K_{\theta}(\phi) - \mu_{\text{target}}(\phi)\right] \right|  +  \sup_{\| \phi \|_{\infty} \leq 1} \left| \sE \left[ \mu_{\theta}^K (\phi) -\hat{\mu}^K_{\theta} (\phi)  \right]  \right|  \\
    & \leq \frac{12 \hat{\rho}}{K} + \beta_2 \\
     & \leq \frac{12 \rho}{K} + \beta_2.
\end{align}

\looseness-1
Where the last inequality follows from the same logic as in~\cref{prop:energybias} where ESS goes up after truncation and therefore $\rho > \hat{\rho}$.
A direct application of Chernoff's inequality gives us $\sP(X < \delta) = \beta_2 \leq \exp(\lambda \delta) \sE[\exp(-\lambda X)]$ where we used the moment generating function of $p_{\theta}(x)$. Thus the additional bias incurred is, 
\begin{equation}
    \sup_{\| \phi \|_{\infty} \leq 1} \left| \sE \left[ \mu_{\theta}^K (\phi) -\mu_{\text{target}} (\phi)  \right] \right| \leq \frac{12 \rho}{K} + \beta_2 \leq \frac{12 \rho}{K} +  \exp(\lambda \delta) \sE[\exp(-\lambda X)].
\end{equation}

Setting $b:= \text{TV}(\mu_{\theta}, \mu_{\text{target}})$ as the bias, then we have
\begin{equation}
   \delta \geq \frac{1}{\lambda} \log \left( \frac{K b}{12 \rho \sE[\exp(-\lambda X)]} \right).
\end{equation}

\end{proof}

\section{Itô Filtering}
\label{sec:proofs}

\subsection{Flow Matching SDE}
\label{app:flow_matching_sde}
As shown in~\citet{domingo2024adjoint} we can write Flow Matching with Gaussian conditional paths and Diffusion models under a unified SDE framework given a reference flow:
\begin{equation}
    x_t = \beta_t x_0 + \alpha_t x_1, 
\end{equation}
where $(\alpha_t)_{t \in [0,1]}, (\beta_t)_{t \in [0,1]}$ are functions such that $\alpha_0 = \beta_1 =0$ and $\alpha_1 = \beta_0 = 1$. In the specific case of flow matching with linear interpolants that we consider we have:
\begin{equation}
    x_t = (1-t) x_0 + t x_1.
\end{equation}
The unified SDE for both flow matching and continuous-time diffusion models as introduced in~\citet{domingo2024adjoint} is then:
\begin{equation}
    dx_t = \kappa_t x + \left( \frac{\sigma_t^2}{2} + \eta_t\right) \mathfrak{s}(x_t, t) + \sigma_t dW_t, \quad \kappa_t = \frac{\dot{\alpha}_t}{\alpha_t}, \eta_t = \beta_t\left( \frac{\dot{\alpha}_t}{\alpha_t}\beta_t - \dot{\beta}_t\right)
\end{equation}
where $\mathfrak{s}(x_t, t)$ is the score function estimated by the diffusion model. Thus the flow matching SDE is:
\begin{equation}
        d x_t = \left(2 f_{t, \theta}(t, x_t) - \frac{x_t}{t}\right) dt + \sigma_t d W_t, \quad \sigma_t = \sqrt{\left(2(1-t)t\right)} 
        \label{eqn:flow_matching_sde}
\end{equation}
In fact, the Stein score can be estimated from the output of a velocity field and vice-versa:
\begin{equation}
    \nabla \log p_t (x_t) = \frac{t f_{t, \theta}(t, x_t) - x_t}{1 - t}, \quad f_{t, \theta}(t, x_t) = \frac{x_t + (1-t)\nabla \log p_t(x_t) }{t}
    \label{eqn:score_as_velocity}
\end{equation}
Rewriting \eqref{eqn:flow_matching_sde} in terms of the score function we get,
\begin{align}
    dx_t &= \frac{x_t}{t} + \sigma^2_t \nabla \log p_t (x_t) + \sigma_t dW_t.
    \label{eqn:flow_matching_sde_with_score}
\end{align}

\subsection{Itô Filtering}
\label{app:ito_filtering}

\vspace{5pt}
\begin{mdframed}[style=MyFrame2]
\begin{restatable}{proposition}{weightsbias}
\label{prop:weights_bias}
\looseness=-1
Assume that the density of the model $p_{\theta}$ after importance sampling $\mu_{\theta}$ is absolutely continuous with respect to the target $\mu_{\text{target}}$. Further, assume that the density of unnormalized importance weights is square integrable $(w(x))^2 < \infty$. Let $r(x_0)$ be the Itô density estimator for $\log p_0(x_0)$ of the flow matching SDE:
\begin{equation}
     dx_t = \frac{x_t}{t} + \sigma^2_t \nabla \mathfrak{s}_{\theta} (t, x_t) + \sigma_t dW_t, \quad \sigma_t =  \sqrt{\left(2(1-t)t\right)}.
\end{equation}
 Given $\rho = 1/ \text{ESS}$, and $\zeta >0$ which is the weight clipping threshold. Then the additional bias of using the It$\hat{\text{o}}$ density estimator for importance sampling $\hat{\mu}_{r, \theta}$ with clipping is:
\begin{equation}
          \sup_{\| \phi \|_{\infty} \leq 1} \left| \sE \left[ \mu_{r, \theta}^K (\phi) -\mu_{\text{target}} (\phi)  \right] \right| \leq \frac{12 \rho}{K} + \beta_3 + \beta_4,
\end{equation}
where $\beta_3 = \text{TV}(\mu_{r, \theta}, \mu_{\theta})$ and $\beta_4 = \text{TV}(\mu_{r, \theta}, \hat{\mu}_{r, \theta})$.

\end{restatable}
\end{mdframed}

We now recall Itô's lemma which states that for a stochastic process,

\begin{equation}
    dx_t = f_t(t, x_t) + g_t dW_t,
\end{equation}
and a smooth function $h: \R \times \R^d \to \R$ the variation of $h$ as a function of the stochastic SDE can be approximated using a Taylor approximation:
\begin{equation}
    d h(t, x_t) = \left ( \frac{\partial}{\partial t} h(t, x_t) + \frac{\partial}{\partial x} h(t, x_t)^T f_t(t, x_t) + \frac{1}{2} \sigma_t^2 \Delta_x h(t, x_t) \right ) dt + \sigma_t \frac{\partial}{\partial x} h(t, x_t) d W_t.
\end{equation}
where $\Delta_x$ is the Laplacian. We will use It\^o's Lemma with $h(t, x_t) := \log p_t(x_t)$ to obtain the It\^o density estimator~\citep{skreta2024superposition,karczewski2024diffusion} but for flow models
\begin{equation}
    d \log p_t(x_t) = \left ( \frac{\partial}{\partial t} \log p_t(x_t) + \frac{\partial}{\partial x} \log p_t(x_t)^T f(t, x_t) + \frac{1}{2} \sigma_t^2 \Delta_x \log p_t(x_t) \right ) dt + \sigma_t \frac{\partial}{\partial x} \log p_t(x_t) d W_t,
\end{equation}
To solve for the change in density over time we can start from the log version of the Fokker-Plank equation:
\begin{equation}
    \frac{\partial}{\partial t} \log p_t(x) = - \nabla \cdot (f(t, x)) + \frac{1}{2} \sigma_t^2 \Delta_x \log p_t(x) - \nabla_x \log p_t(x)^T \left ( f(t, x) - \frac{1}{2} \sigma_t^2 \nabla_x \log p_t(x) \right )
\end{equation}

in the general case we end with:
\begin{equation}\label{eq:general}
    d \log p_t(x_t) = \left ( - \nabla \cdot \left ( f(t, x_t) - \sigma_t^2 \nabla_x \log p_t(x_t) \right ) + \frac{1}{2} \sigma_t^2 \| \nabla_x \log p_t(x_t) \|^2 \right ) dt + \sigma_t \nabla_x \log p_t(x_t)^T d W_t.
\end{equation}

We now apply this to the flow-matching SDE~\eqref{eqn:flow_matching_sde_with_score} written in terms of the score function. In particular, we have,
\begin{align}
    d \log p_t(x_t) &= \left ( - \nabla \cdot \left (\sigma^2_t \nabla_x \log p_t(x_t) + \frac{x_t}{t} - \sigma_t^2 \nabla_x \log p_t(x_t) \right ) + \frac{1}{2} \sigma_t^2 \| \nabla_x \log p_t(x_t) \|^2 \right ) dt  \nonumber\\
    & + \sigma_t \nabla_x \log p_t(x_t)^T d W_t \nonumber \\
    d \log p_t(x_t) &= \left ( - d / t + \frac{1}{2} \sigma_t^2 \| \nabla_x \log p_t(x_t) \|^2 \right ) dt + \sigma_t \nabla_x \log p_t(x_t)^T d W_t.
    \label{eqn:flow_matching_ito_density}
\end{align}
The above equation makes an implicit assumption that we have access to the actual ground truth score function of $\nabla \log_t(x_t)$ rather than the estimated one $\mathfrak{s}_{\theta}$, expressed via the vector field as in~\eqref{eqn:score_as_velocity}. When working with imperfect score estimates we have the following SDE:
\begin{align}
    dx_t &= \frac{x_t}{t} + \sigma^2_t \nabla \mathfrak{s}_{\theta} (t, x_t) + \sigma_t dW_t.
    \label{eqn:flow_matching_sde_with_score_estimate}
\end{align}

The score estimation error causes a discrepancy in $\log p_t(x_t)$ estimates
whose error is captured in the theorem from~\citet{karczewski2024diffusion}[Theorem 3]:

\begin{equation}
   \log r_0(x_0) = \log p_0(x_0)  + Y
\end{equation}
where $\log r_0$ is the bias of the log density starting at time $t=0$ of the auxiliary process that does not track $x_t$ correctly due to the estimation error of the score. Also, $Y$ is a random variable such that that bias of $r_0$ is given by:

\begin{equation}
    \sE[Y] = \underbrace{\frac{1}{2}\sE_{t\sim \gU(0,1), x_t \sim p_t(x_t)} \left[ \sigma^2_t||\mathfrak{s}_{\theta}(t, x_t) - \nabla \log p_t(x_t) ||^2 \right] }_{\geq 0}
\end{equation}

Thus the Itô density estimator forms an upper bound to the true log density, i.e. $r_0 (x_0)\geq \log p_0 (x_0)$. This allows us to form an upper bound on the normalized log weights as an expectation,
\begin{align*}
    \sE_{x_0 \sim p_{\theta}(x_0)}[\log \bar{w}(x_0)] &=  \sE_{x_0 \sim p_{\theta}(x_0)} \left[-\frac{\gE(x_0)}{k_BT} - \log p_0 (x_0) - C \right]\\
                & \leq \sE_{x_0 \sim p_{\theta}(x_0)} \left[-\frac{\gE(x_0)}{k_BT} - r_0 (x_0) \right],
\end{align*}
where $C$ is a constant. We define $ \log \bar{w}_r(x_0):= -\frac{\gE(x_0)}{k_BT} - r_0 (x_0)$ as the new normalized importance weights, module constants. 
We can now compute the additional bias of self-normalized importance sampling estimator $\mu^K_{r, \theta}$
\begin{align}
    \text{TV}(\mu_{r, \theta}, \mu_{\theta}) &=  \sup_{\| \phi \|_{\infty} \leq 1} \left| \sE \left[ \mu_{r, \theta}^K (\phi) - \mu_{\theta}^K (\phi)  \right] \right| \\
    &=   \sup_{\| \phi \|_{\infty} \leq 1} \left| \sE_{x^i \sim p_{\theta}} \left[ \sum^K_{i=1}\bar{w}_r(x^i) \phi(x^i) \right]  - \sE_{x^j \sim p_{\theta}} \left[\sum^K_{j=1} \bar{w}(x^j)\phi(x^j)  \right] \right| \\
   & = \frac{1}{2}\left( \sE_{x^i \sim p_{\theta}} \left[\sum^K_{i=1} \bar{w}_r(x^i) \right]  - \sE_{x^j \sim p_{\theta}} \left[ \sum^K_{j=1} \bar{w}(x^j)  \right] \right) \\
   &= \frac{1}{2}\left( \sE_{x^i \sim p_{\theta}} \left[\sum^K_{i=1} \exp\left( \frac{1}{2}\sE_{t\sim \gU(0,1), x_t \sim p_t(x_t)} \left[ \sigma^2_t||\mathfrak{s}_{\theta}(t, x_t) - \nabla \log p_t(x_t) ||^2 \right] \right)  \right] \right) \\
   & := \beta_3
\end{align}

The total bias is then

\begin{equation}
    \sup_{\| \phi \|_{\infty} \leq 1} \left| \sE \left[ \mu_{r, \theta}^K (\phi) -\mu_{\text{target}} (\phi)  \right] \right| \leq \frac{12 \rho}{K} + \beta_3 .
\end{equation}
Finally, when clipping weights with $\zeta > 0 $ we induce a truncated distribution $\hat{\mu}_{r, \theta}$, i.e. $\hat{r}_0:=\sP(r_0{x_0} > \zeta)$. Using~\cref{lemma:tvd_truncation} this creates another constant factor that contributes $\text{TV}(\mu_{r, \theta}, \hat{\mu}_{r, \theta}) =\beta_4$ to the overall bias:
\begin{equation}
      \sup_{\| \phi \|_{\infty} \leq 1} \left| \sE \left[ \mu_{r, \theta}^K (\phi) -\mu_{\text{target}} (\phi)  \right] \right| \leq \frac{12 \rho}{K} + \beta_3 + \beta_4.
\end{equation}

\section{Datasets}
\label{app:dataset_details}

\looseness=-1
For all datasets besides alanine dipeptide we use a training set of \scione{5} contiguous samples (\SI{1}{\us} simulation time) from a single MCMC chain, a validation set of the next \sci{2}{4} contiguous samples (\SI{0.2}{\us} simulation time), and a test set of \scione{4} uniformly strided subsamples from the remaining trajectory. Since these are highly multimodal energy functions, this leaves us with \text{biased} training data relative to the Boltzmann distribution; we split trajectories this way to test the model in a challenging and realistic setting. We describe the datasets below and present the simulation parameters in~\cref{tab:datasets}. Ramachandran plots for the training and test data (before subsampling) are provided in \cref{fig:rama-al23,fig:rama-al3,fig:rama-al4,fig:rama-al6}.

\begin{table}[h]
    \centering
    \caption{Overview of molecular dynamics simulation parameters.}
    \label{tab:datasets}
    \begin{tabular}{l r r r}
        \toprule
        Peptide & Force field & Temperature & Time step \\
        \midrule
        Alanine dipeptide & Amber ff99SBildn & $300\text{K}$ & $1\text{fs}$ \\
        Trialanine & Amber 14 & $310\text{K}$ & $1\text{fs}$ \\
        Alanine tetrapeptide & Amber ff99SBildn & $300\text{K}$ & $1\text{fs}$ \\
        Hexaalanine & Amber 14 & $310\text{K}$ & $1\text{fs}$ \\
        Chignolin & Amber 14 & $310\text{K}$ & $1\text{fs}$ \\
        \bottomrule
    \end{tabular}
\end{table}

\xhdr{Alanine dipeptide} For this dataset we use the data and data split from~\citet{klein2024transferable}. Here the training set is purposely biased with an overrepresentation of an underrepresented mode, i.e. the positive $\varphi$ state. This  bias makes it easier to reweight to the target Boltzmann distribution.
Alanine Dipeptide consist of one Alanine amino acids, an acetyl group, and an N-methyl group.

\xhdr{Trialanine and hexa-alanine}
For the peptides composed of multiple alanine amino acids, we generate MD trajectories using the \textit{OpenMM} library \citep{eastman2017openmm}. All simulations are conducted in implicit solvent, with the simulation parameters detailed in~\cref{tab:datasets}. These systems do not include any additional capping groups, such as those present in alanine dipeptide and alanine tetrapeptide, as they are generated in the same manner as described in \citet{klein2023timewarp}. There are two peptide bonds in trialanine and five in hexa-alanine, resulting in two and five Ramachandran plots respectively.

\xhdr{Alanine Tetrapeptide  (AD4)} 
For this dataset we use the same system setup as in \citet{dibak2021temperature}, but treat all bonds as flexible. The original dataset kept all hydrogen bonds fixed, as the Boltzmann Generator was operating in internal coordinates. The MD simulation to generate the dataset is then performed as described above. Alanine Tetrapeptide consist of three Alanine amino acids, an acetyl group, and an N-methyl group. Therefore, there are four Ramachandran plots.

\xhdr{Chignolin} 
In addition to the small peptide systems, we also investigate the small protein chignolin, consisting of ten amino acids (\texttt{GYDPETGTWG}). We simulate this system using the same configuration as trialanine, defined in \cref{tab:datasets} for \SI{4.5}{\us}.

\begin{figure}[H]
    \centering
    \captionsetup[subfigure]{aboveskip=-1pt,belowskip=-1pt}
    \begin{subfigure}{0.24\linewidth}
    \end{subfigure}
    \begin{subfigure}{0.24\linewidth}
            \includegraphics{new_media/rama_train_al2_0.pdf}
            \caption{Train}
    \end{subfigure}
    \begin{subfigure}{0.24\linewidth}
            \includegraphics{new_media/rama_test_al2_0.pdf}
            \caption{Test}
    \end{subfigure}
    \caption{Alanine dipeptide Ramachandran plots for train and test data subsets. The lower probability state is oversampled in the training data as in \citet{klein2024transferable}.}
    \label{fig:rama-al23}
\end{figure}

\begin{figure}[H]
    \centering
    \captionsetup[subfigure]{aboveskip=-1pt,belowskip=-1pt}
    \begin{subfigure}{0.49\linewidth}
    \centering
        \begin{subfigure}{0.49\linewidth}
            \includegraphics{new_media/rama_train_al3_0.pdf}
        \end{subfigure}
        \begin{subfigure}{0.49\linewidth}
            \includegraphics{new_media/rama_train_al3_1.pdf}
        \end{subfigure}
    \caption{Train}
    \end{subfigure}
    \begin{subfigure}{0.49\linewidth}
    \centering
        \begin{subfigure}{0.49\linewidth}
            \includegraphics{new_media/rama_test_al3_0.pdf}
        \end{subfigure}
        \begin{subfigure}{0.49\linewidth}
            \includegraphics{new_media/rama_test_al3_1.pdf}
        \end{subfigure}
    \caption{Test}
    \end{subfigure}
    \caption{Trialanine Ramachandran plots for train and test data. We observe a missing mode in the first Ramachandran plot of the training subset.}
    \label{fig:rama-al3}
\end{figure}

\begin{figure}[H]
    \centering
    \captionsetup[subfigure]{aboveskip=-1pt,belowskip=-1pt}
    \begin{subfigure}{\linewidth}
    \centering
        \begin{subfigure}{0.19\linewidth}
            \includegraphics{new_media/rama_train_al4_0.pdf}
        \end{subfigure}
        \begin{subfigure}{0.19\linewidth}
            \includegraphics{new_media/rama_train_al4_1.pdf}
        \end{subfigure}
        \begin{subfigure}{0.19\linewidth}
            \includegraphics{new_media/rama_train_al4_2.pdf}
        \end{subfigure}
    \caption{Train}
    \end{subfigure}
    \begin{subfigure}{\linewidth}
    \centering
        \begin{subfigure}{0.19\linewidth}
            \includegraphics{new_media/rama_test_al4_0.pdf}
        \end{subfigure}
        \begin{subfigure}{0.19\linewidth}
            \includegraphics{new_media/rama_test_al4_1.pdf}
        \end{subfigure}
        \begin{subfigure}{0.19\linewidth}
            \includegraphics{new_media/rama_test_al4_2.pdf}
        \end{subfigure}
    \caption{Test}
    \end{subfigure}
    \caption{Alanine tetrapeptide Ramachandran plots for train and test data subsets. We obverse an underrepresented mode in the second Ramachandran plot of the training subset.}
    \label{fig:rama-al4}
\end{figure}

\begin{figure}[H]
    \centering
    \captionsetup[subfigure]{aboveskip=-1pt,belowskip=-1pt}
    \begin{subfigure}{\linewidth}
    \centering
        \begin{subfigure}{0.19\linewidth}
            \includegraphics{new_media/rama_train_al6_0.pdf}
        \end{subfigure}
        \begin{subfigure}{0.19\linewidth}
            \includegraphics{new_media/rama_train_al6_1.pdf}
        \end{subfigure}
        \begin{subfigure}{0.19\linewidth}
            \includegraphics{new_media/rama_train_al6_2.pdf}
        \end{subfigure}
        \begin{subfigure}{0.19\linewidth}
            \includegraphics{new_media/rama_train_al6_3.pdf}
        \end{subfigure}
        \begin{subfigure}{0.19\linewidth}
            \includegraphics{new_media/rama_train_al6_4.pdf}
        \end{subfigure}
    \caption{Train}
    \end{subfigure}
    \begin{subfigure}{\linewidth}
    \centering
        \begin{subfigure}{0.19\linewidth}
            \includegraphics{new_media/rama_test_al6_0.pdf}
        \end{subfigure}
        \begin{subfigure}{0.19\linewidth}
            \includegraphics{new_media/rama_test_al6_1.pdf}
        \end{subfigure}
        \begin{subfigure}{0.19\linewidth}
            \includegraphics{new_media/rama_test_al6_2.pdf}
        \end{subfigure}
        \begin{subfigure}{0.19\linewidth}
            \includegraphics{new_media/rama_test_al6_3.pdf}
        \end{subfigure}
        \begin{subfigure}{0.19\linewidth}
            \includegraphics{new_media/rama_test_al6_4.pdf}
        \end{subfigure}
    \caption{Test}
    \end{subfigure}
    \caption{Hexa-alanine Ramachandran plots for train and test data subsets. We observe good mode coverage of the training data.}
    \label{fig:rama-al6}
\end{figure}

\section{Experimental Details}\label{app:exp_details}

\subsection{Metrics}

For computational efficiency we subsample \scione{4} reference samples from the evaluation trajectory to serve as ground truth. Similarly in cases where a method produces more than \scione{4} samples, a random subset of size \scione{4} is selected for comparison. We quantify distributional similarity using empirical Wasserstein-2 distances between generated samples and the reference data. Given two empirical distributions, $\mu = \frac{1}{n} \sum_{i=1}^n \delta_{x_i}$ and $\nu = \frac{1}{m} \sum_{j=1}^m \delta_{y_j}$, the Wasserstein-2 distance is computed as

\begin{equation}
W_2(\mu, \nu) = \min_{\pi \in \Pi(\mu, \nu)} \sqrt{\sum_{i=1}^{n} \sum_{j=1}^{m} \pi_{ij} \, c(x_i, y_j)^2},
\end{equation}

where $\Pi(\mu, \nu)$ denotes the set of admissible transport plans and $c(x, y)^2$ is a defined cost function. Optimal couplings are computed using the \texttt{POT} library \citep{flamary2021pot}.

\xhdr{Energy cost} To assess energy distribution similarity, we define the following cost

\begin{equation}
c_E(x, y)^2 = \left| E(x) - E(y) \right|^2.
\end{equation}

\xhdr{Dihedral torus cost} We evaluate macrostructural similarity in the space of backbone dihedral angles $(\phi, \psi)$, which encode conformational information. For a molecule with $L$ residues, we define the angle vector:

\begin{equation}
\operatorname{Dihedrals}(x) = (\phi_1, \psi_1, \ldots, \phi_{L-1}, \psi_{L-1}).
\end{equation}

Due to angle periodicity, we define the cost on the resulting torus as:

\begin{equation}
c_\mathbb{T}(x, y)^2 = \sum_{i=1}^{2L} \left[ (\operatorname{Dihedrals}(x)_i - \operatorname{Dihedrals}(y)_i + \pi) \bmod 2\pi - \pi \right]^2,
\end{equation}

capturing angular deviations while respecting circular geometry.

\paragraph{TICA projection cost}

Time-lagged Independent Component Analysis (TICA) performs dimensionality reduction for identification of slow dynamical modes. From the mean-centered time series $\tilde{x}_t$, we compute:

\begin{equation}
\hat{C}_{00} = \frac{1}{T-\tau}\sum_{t=1}^{T-\tau} \tilde{x}_t \tilde{x}_t^\top, \quad \hat{C}_{0\tau} = \frac{1}{T-\tau}\sum_{t=1}^{T-\tau} \tilde{x}_t \tilde{x}_{t+\tau}^\top,
\end{equation}

and solve the generalized eigenvalue problem:

\begin{equation}
C_{0\tau} w = \lambda C_{00} w.
\end{equation}

The top two eigenvectors $w_1$ and $w_2$ define projections capturing the most autocorrelated directions. Using these, we define the TICA cost:

\begin{equation}
c_\text{TICA}(x, y)^2 = \sum_{j=1}^2 \left[ w_j^\top x - w_j^\top y \right]^2.
\end{equation}

Note that the TICA basis is computed from the full evaluation trajectory (un-subsampled), while the comparison subset is restricted to the \scione{4} subset. TICA analysis is performed only on heavy atom coordinates.

\subsection{Timings}

\xhdr{Sampling time calculations} For the sampling inference times in \cref{fig:computational_cost_in_hours}, we compute all times on a single NVIDIA L40S GPU, using the maximum power of two batch size possible.

\xhdr{Training time}
For training times we compute all times on a single A100 80GB GPU except for $\sethree$-EACF which is trained on a single H100. We report the total time in hours until convergence for all methods in the table below.

\begin{table}[h]
    \centering
    \caption{Training time (hours) for all methods.}
    \begin{tabular}{cccccc}
        \hline
        Model & ALDP & AL3 & AL4 & AL6 & Chignolin \\ 
        \hline
        $\sethree$-EACF & 160 & --- & ---& --- & ---\\
        ECNF++ & 9.72 & 12.5&  17.17 &  76.94 & --- \\
        \nameshort & 16.83 & 24.67 & 41.67 & 57.5 & 427.33 \\ 
        \hline
    \end{tabular}
    \label{tab:training_times}
\end{table}

\subsection{$\sethree$-EACF Implementation Details}
\label{app:eacf_implementation_details}
\xhdr{Equivariant augmented coupling flow (EACF)~\citep{midgley2023se}} We adopt the original model configuration from ~\citep{midgley2023se} for our EACF baseline on ALDP. Specifically, we choose the more stable spherical-projection EACF with a 20-layer configuration. Each layer has two ShiftCoM layers and two core-transformation blocks. The EGNN used in the core transformation block consists of 3 message-passing layers with 128 hidden states. Stability enhancement tricks like stable MLP and dynamic weight clipping on each layer's output are fully applied. The model is trained for 50 epochs with a bathh size of 20 using Adam optimizer and peak learning rate of \(1\times10^{-4}\). We use the default 20 samples for likelihood estimation.

\xhdr{EACF as a Boltzmann generator} EACF leverages augmented dimensions, and therefore to estimate the likelihood of a sample $x$ under an EACF model, we need to use an estimate based on samples from the augmented dimension $a$. Specifically, for a Gaussian distributed augmented variable $a$, we can estimate the marginal density of an observation as
\begin{equation}
    q(x) = \mathbb{E}_{a \sim \pi(\cdot | x)} \left [ \frac{q(x,a)}{\pi(a | x)} \right ],
\end{equation}
however, this is only a consistent estimator of the likelihood and for finite sample sizes has variance. This makes EACF unsuitable for our application of large-scale Boltzmann generators, as in this setting we need to compute exact likelihoods. Variance in likelihood estimation would lead to bias in the final distribution under self-normalized importance sampling or a \nameshort strategy. We therefore do not consider EACF as a viable option for large scale Boltzmann distribution sampling.

\subsection{ECNF Implementation Details}
\label{app:ecnf_implementation_details}

\subsubsection{Network and Training}
\begin{algorithm}[H]
  \caption{ECNF flow matching training}
  \label{alg:cfm}
\begin{algorithmic}
\STATE {\bfseries Input:} Prior $q_0$, Empirical samples from data $q_1$, bandwidth $\sigma$, batchsize $b$, initial network $v_{\theta}$.
\WHILE{Training}
\STATE $\vx_0 \sim q_0(\vx_0); \quad \vx_1 \sim q_1(\vx_1)$ \COMMENT{Sample batches of size $b$ \textit{i.i.d.} from the dataset}
\STATE $t \sim \mathcal{U}(0, 1)$
\STATE $\mu_t \gets t \vx_1 + (1 - t) \vx_0$
\STATE $x \sim \mathcal{N}(\mu_t, \sigma^2 I)$
\STATE $\mathcal{L}(\theta) \gets \| v_\theta(t, x) - (\vx_1 - \vx_0)\|^2$
\STATE $\theta \gets \mathrm{Update}(\theta, \nabla_\theta \mathcal{L}(\theta))$
\ENDWHILE
\STATE \textbf{Return} $v_\theta$
\end{algorithmic}
\end{algorithm}

\xhdr{Equivariant continuous normalizing flow (ECNF)~\citep{klein2024transferable}} We use the supplied pretrained model from \citet{klein2024transferable} for our ECNF baseline on alanine dipeptide. Therefore all training parameters are equivalent to, and specified in, that work. We use the specification for the model ``TBG + Full'' in that work. 

\xhdr{ECNF++} We note four improvements to the ECNF, which together substantially improve scalability.
\begin{enumerate}
    \item \textbf{Flow matching loss.} In \citet{klein2024transferable} a flow matching algorithm with smoothing is employed which provides extra stability during training. This is depicted in Alg.~\ref{alg:cfm}, however this smooths out the optimal target distribution~\citep[Proposition 3.3]{tong_conditional_2023}. ECNF uses $\sigma = 0.01$ where we use $\sigma=0$ for ECNF++. We find that $\sigma > 0$ in this case causes poor molecular structures to be generated as the bond lengths are not able to be controlled precisely enough. We note that $\sigma=0$ is used in most recent large scale flow matching models~\cite{liu2024instaflowstephighqualitydiffusionbased,esser2024scalingrectifiedflowtransformers}.
    \item \textbf{Architecture size.} Empirically, we find the ECNF to be underparameterized. We perform a grid search over layer width and depth, finding a width of 256 and depth of 5 block to be a good balance between performance and speed on alanine dipeptide. We employ the same parameters for larger molecular systems.
    \item \textbf{Improved optimizer and LR scheduler.} We find using an AdamW \citep{loshchilov2017decoupled} with moderate weight decay of \scione{-4} improves performance and stability. Prior work has found weight decay helps to keep the Lipschitz constant of the flow low and avoids stiff dynamics which enables accurate ODE solving during inference. We also use a smoothly varying cosine schedule with warm-up (over 5\% of iteration budget) which enables a larger maximum learning rate and faster training than the two step schedule used previously. Both the start and end learning rates are 500 times lower than the defined maximum.
    \item \textbf{Exponential moving average.} We use an exponential moving average (EMA) on the weights with decay $0.999$. This is standard practice in flow models, which improves performance.
\end{enumerate}
These four elements together greatly improve the ECNF training, enabling larger systems to be successfully modeled, and provide a strong foundation for future Boltzmann generator training on molecular systems using equivariant continuous normalizing flows. Qualitatively, we find ECNFs quite stable to train and robust to training parameters relative to invertible architectures. However, it is very slow to compute the exact likelihoods necessary for importance sampling.

\xhdr{Other parameters} For inference we use a Dormand-Prince 45 (dopri5) adaptive step size solver with absolute tolerance $10^{-4}$ and relative tolerance $10^{-4}$.

\xhdr{Likelihood evaluation} Evaluating the likelihood of a CNF model requires calculating the integral of the divergence, as in~\eqref{eqn:instantaneous_change_of_variable}. While there exist fast unbiased approximations of the likelihood using Hutchinson's trace estimator~\cite{Hutchinson01011990,grathwohl_ffjord_2018}, these are unfortunately unsuitable for Boltzmann generator applications where variance in the likelihood estimator leads to biased weights under self-normalized importance sampling. We therefore calculate the Jacobian using automatic differentiation which is both memory and time intensive. For example, on hexa-alanine, the maximum batch size that can fit on an 80GB A100 GPU is 8. This batch takes around 2 minutes for 84 integration steps. We also use an improved vectorized Jacobian trace implementation for all CNF which reduces memory by roughly half and time by roughly 3x over the previous implementation \citep{klein2024transferable}.

\xhdr{On using a CNF proposal with \nameshort} In principle it is possible to drop in replace our NF architecture with a CNF in \nameshort. However, there are several drawbacks to such an approach, most notably in efficiency. As previously discussed, CNFs are extremely computationally inefficient to sample a likelihood from. We find on the order of 100 SBG steps are necessary for best performance. This would make CNFs at least two orders of magnitude slower to sample from, when they are already at the edge of tractability for the current importance sampling estimates. We leave it to future work to consider faster CNF likelihoods and note that our \nameshort algorithm could be applied there readily.

\subsection{SBG Implementation Details}
\label{app:abgen_implementation_details}

\xhdr{Architecture} We scale the TarFlow architecture for increasingly challenging datasets. As advised by \citet{zhai2024normalizing} we scale the layers per block alongside the number of blocks. The layers / blocks / channels and resulting number of parameters are presented in Table \ref{tab:model_configs}. We note the larger number of parameters in the TarFlow relative to the ECNF++ despite the faster inference walltime, due to the lack of simulation and higher computational efficiency of the architecture.

\begin{table}[h]
    \caption{TarFlow configurations across different datasets.}
    \label{tab:model_configs}
    \centering
    \begin{tabular}{lcccc}
        \toprule
        Dataset & Layers per Block & Number Blocks & Channels & Number Parameters (M) \\
        \midrule
        ALDP & 4 & 4 & 256 & 13 \\
        AL3 & 6 & 6 & 256 & 29 \\
        AL4 & 6 & 6 & 384 & 64 \\
        AL6 & 6 & 6 & 384 & 64 \\
        Chignolin & 8 & 8 & 384 & 114 \\
        \bottomrule
    \end{tabular}
\end{table}

\xhdr{Training configuration} The training hyperparameters used closely follow those of \citet{zhai2024normalizing}, although we deviate in using a larger value of weight decay as instability was observed during training. Namely we use a learning rate of \(1 \times 10^{-4}\), weight decay of \(4 \times 10^{-4}\), Adam \(\beta_1,\beta_2\) of \((0.90, 0.95)\). We additionally employ the same cosine decay learning rate schedule with warmup (start and end learning rate 500 times lower than maximum value) and exponential moving average decay (0.999) used in ECNF++. Training is performed for 1000 epochs. Center of mass augmentation is applied with a standard deviation of \(\frac{1}{\sqrt{n}}\), for \(n\) the number of particles, to match that of the prior for a given system. As non-monotonic improvement was observed on validation metrics, we use early stopping on the SNIS \emetric against the validation dataset.

\begin{table}
    \centering
    \caption{Overview of training configurations}
    \label{tab:training}
    \begin{tabular}{l r r r}
        \toprule
        Training Parameter & ECNF & ECNF++ & TarFlow\\
        \midrule
        Learning Rate & $5 \times 10^{-4}$ & $5 \times 10^{-4}$ & $1 \times 10^{-4}$\\
        Weight Decay & 0.0 &$1 \times 10^{-2} $& $4 \times 10^{-4}$\\
        $\beta_1,\beta_2$ &0.9, 0.999 & 0.9, 0.999 & 0.9, 0.95\\
        EMA Decay & 0.000 & 0.999 & 0.999\\
        Width & 64 & 256 & Varies\\
        $N$ blocks & 5 & 5 & Varies\\
        Parameters & 152 K & 2.317 M & Varies  \\
        \bottomrule
    \end{tabular}
\end{table}

\xhdr{Sampling hyperparameters} Whilst the TarFlow is capable of generating low-energy peptide states, it is also prone to generating samples of extremely high target energy. For standard importance sampling this presents no issue as these samples will be assigned negligible importance weight. However, for the \nameshort these high energy samples were prone to numerical instability during Langevin dynamics. To mitigate this issue we truncate the proposal distribution prior \(p_\theta\) based on an energy cutoff, noting that the bias introduced by this operation is bounded in \cref{prop:energybias}.
Similarly to EACF, we additionally remove the samples corresponding to the largest 0.2\% of importance weights, in the case of \nameshort this is performed once, prior to Langevin dynamics. See \S\ref{sec:ablations} for an ablation on the effect of weight clipping in \nameshort. For alanine systems we use 100 Langevin time steps, with \(\text{ESS}_\text{threshold} = 0.5\) and \(\eps = 1\times10^{-5}\) up to trialanine, and \(\eps = 1\times10^{-6}\) thereafter. For chignolin we use 500 time steps, with \(\text{ESS}_\text{threshold} = 0.5\) and \(\eps = 1\times10^{-5}\). For ablations of these hyperparameters see \S\ref{sec:ablations}.

\section{Additional Results}
\label{app:additional_results}

\subsection{Chirality}

The ECNF architecture is E(3) equivariant, hence is equivariant to reflections, and will generate samples of both possible global chiralities. As the energy functions are themselves invariant to reflections this is not resolved by importance sampling. Having the correct global chirality is necessary to match the test dataset dihedral angle distributions where only one global chirality is present in the data. The incorrect chirality can show up as a symmetric mode on Ramachandran plots. To resolve this issue we follow \citet{klein2024transferable} in detecting incorrect chirality conformations, and reflecting them. However, unlike \citet{klein2024transferable} points with unresolvable symmetry (e.g mixed chirality conformations) \emph{are not} dropped. The results for \torusmetric before and after fixed-chirality samples are presented in Table \ref{tab:chirality_results}. We observe a reduction in metric value (improved performance) on all configurations, which we attribute to evaluation noise. We further note that non-equivariant methods such as \nameshort do not suffer this effect and hence do not require any symmetry post-processing.

\begin{table*}[ht!]
\caption{$\sT$-$\gW_2$ results for unprocessed and fixed-chirality samples from ECNF and ECNF++}
\label{tab:chirality_results}
\begin{tabular}{@{}lcccccc}
    \toprule
    Datasets $\rightarrow$ & \multicolumn2c{Tripeptide (AL3)} & \multicolumn2c{Tetrapeptide (AL4)} & \multicolumn2c{Hexapeptide (AL6)}  \\
    \cmidrule(lr){2-3}\cmidrule(lr){4-5}\cmidrule(lr){6-7}
    Algorithm $\downarrow$ & Unprocessed & Fixed & Unprocessed & Fixed & Unprocessed & Fixed \\
    \midrule
    ECNF++ (Ours)   
         & 1.967 $\pm$ 0.062 & 1.177 $\pm$ 0.145 
         & 2.414 $\pm$ 0.000 & 2.082 $\pm$ 0.005
         & 5.405 $\pm$ 0.069 & 4.315 $\pm$ 0.018 \\
    \bottomrule
\end{tabular}
\vspace{-5pt}
\end{table*}

\subsection{Ramachandran Plots}
\label{app:rama_plots}

In this appendix we include the Ramachandran plots for each model on each peptide system. Please note that the ground truth training and test Ramachandran plots are presented in~\S\ref{app:dataset_details}.

\xhdr{Alanine dipeptide} In Figure~\ref{fig:ramas_aldp} we present the alanine dipetide Ramachandran plots. Both ECNF and \nameshort cover all relevant modes. We find that that ECNF++ models the distribution well, but drops the positive \(\varphi\) mode. This is notable as this mode is oversampled in the training data (see Figure~\ref{fig:rama-al23}).

\xhdr{Trialanine} In Figure~\ref{fig:ramas_al3} we can see the Ramachandran plot for resampled points for the trialanine dataset. Comparing this to the train and test data in Figure~\ref{fig:rama-al3} we see that the AL3 training data is missing the \(\varphi_1\) positive mode which is reflected in all of the models. %

\xhdr{Alanine tetrapeptide} In Figure~\ref{fig:ramas_al4} we present Ramachandran plots for resampled points on the alanine tetrapeptide dataset. Comparing this to the train and test distributions in Figure \ref{fig:rama-al4} we observe that both ECNF++ and \nameshort capture the dihedral angle distribution with comparable success.

\xhdr{Hexa-alanine} In Figure~\ref{fig:ramas_al6} we present the Ramchandran plots for samples from ECNF++ and \nameshort. We find that \nameshort succeeds to capture the low density positive \(\varphi\) modes, albeit with a tight concentration of points as opposed to a broad range of low density. We additionally observe the negative \(\Psi_1\) mode to be well captured by the \nameshort.

\begin{figure}[H]
    \centering
    \captionsetup[subfigure]{aboveskip=-1pt,belowskip=-1pt}
    \begin{subfigure}{0.24\linewidth}
            \includegraphics[scale=0.95]{new_media/rama_eacf_al2_0.pdf}
            \caption{EACF}
    \end{subfigure}
    \begin{subfigure}{0.24\linewidth}
            \includegraphics[scale=0.95]{new_media/rama_ecnf_al2_0.pdf}
            \caption{ECNF}
    \end{subfigure}
    \begin{subfigure}{0.24\linewidth}
            \includegraphics[scale=0.95]{new_media/rama_ecnf++_al2_0.pdf}
            \caption{ECNF++}
    \end{subfigure}
    \begin{subfigure}{0.24\linewidth}
            \includegraphics[scale=0.95]{new_media/rama_tarflow_al2_0.pdf}
            \caption{\nameshort}
    \end{subfigure}
    \caption{Alanine dipeptide Ramachandran plots for baseline methods (SNIS) and \nameshort (SMC). \scione{4} points sampled per method.}
    \label{fig:ramas_aldp}
\end{figure}

\begin{figure}[H]
    \centering
    \captionsetup[subfigure]{aboveskip=-1pt,belowskip=-1pt}
    \begin{subfigure}{0.49\linewidth}
    \centering
        \begin{subfigure}{0.49\linewidth}
            \includegraphics[scale=0.95]{new_media/rama_ecnf++_al3_0.pdf}
        \end{subfigure}
        \begin{subfigure}{0.49\linewidth}
            \includegraphics[scale=0.95]{new_media/rama_ecnf++_al3_1.pdf}
        \end{subfigure}
    \caption{ECNF++}
    \end{subfigure}
    \begin{subfigure}{0.49\linewidth}
    \centering
        \begin{subfigure}{0.49\linewidth}
            \includegraphics[scale=0.95]{new_media/rama_tarflow_al3_0.pdf}
        \end{subfigure}
        \begin{subfigure}{0.49\linewidth}
            \includegraphics[scale=0.95]{new_media/rama_tarflow_al3_1.pdf}
        \end{subfigure}
    \caption{\nameshort}
    \end{subfigure}
    \caption{Trialanine Ramachandran plots for ECNF++ (SNIS) and \nameshort (SMC). \scione{4} points sampled per method.}
    \label{fig:ramas_al3}
\end{figure}

\begin{figure}[H]
    \centering
    \captionsetup[subfigure]{aboveskip=-1pt,belowskip=-1pt}
    \begin{subfigure}{\linewidth}
    \centering
        \begin{subfigure}{0.19\linewidth}
            \includegraphics[scale=0.95]{new_media/rama_ecnf++_al4_0.pdf}
        \end{subfigure}
        \begin{subfigure}{0.19\linewidth}
            \includegraphics[scale=0.95]{new_media/rama_ecnf++_al4_1.pdf}
        \end{subfigure}
        \begin{subfigure}{0.19\linewidth}
            \includegraphics[scale=0.95]{new_media/rama_ecnf++_al4_2.pdf}
        \end{subfigure}
    \caption{ECNF++}
    \end{subfigure}
    \begin{subfigure}{\linewidth}
    \centering
        \begin{subfigure}{0.19\linewidth}
            \includegraphics[scale=0.95]{new_media/rama_tarflow_al4_0.pdf}
        \end{subfigure}
        \begin{subfigure}{0.19\linewidth}
            \includegraphics[scale=0.95]{new_media/rama_tarflow_al4_1.pdf}
        \end{subfigure}
        \begin{subfigure}{0.19\linewidth}
            \includegraphics[scale=0.95]{new_media/rama_tarflow_al4_2.pdf}
        \end{subfigure}
    \caption{\nameshort}
    \end{subfigure}
    \caption{Alanine tetrapeptide Ramachandran plots for ECNF++ (SNIS) and \nameshort (SMC). \scione{4} points sampled per method.}
    \label{fig:ramas_al4}
\end{figure}

\begin{figure}[H]
    \centering
    \captionsetup[subfigure]{aboveskip=-1pt,belowskip=-1pt}
    \begin{subfigure}{\linewidth}
    \centering
        \begin{subfigure}{0.19\linewidth}
            \includegraphics[scale=0.95]{new_media/rama_ecnf++_al6_0.pdf}
        \end{subfigure}
        \begin{subfigure}{0.19\linewidth}
            \includegraphics[scale=0.95]{new_media/rama_ecnf++_al6_1.pdf}
        \end{subfigure}
        \begin{subfigure}{0.19\linewidth}
            \includegraphics[scale=0.95]{new_media/rama_ecnf++_al6_2.pdf}
        \end{subfigure}
        \begin{subfigure}{0.19\linewidth}
            \includegraphics[scale=0.95]{new_media/rama_ecnf++_al6_3.pdf}
        \end{subfigure}
        \begin{subfigure}{0.19\linewidth}
            \includegraphics[scale=0.95]{new_media/rama_ecnf++_al6_4.pdf}
        \end{subfigure}
    \caption{ECNF++}
    \end{subfigure}
    \begin{subfigure}{\linewidth}
    \centering
        \begin{subfigure}{0.19\linewidth}
            \includegraphics[scale=0.95]{new_media/rama_tarflow_al6_0.pdf}
        \end{subfigure}
        \begin{subfigure}{0.19\linewidth}
            \includegraphics[scale=0.95]{new_media/rama_tarflow_al6_1.pdf}
        \end{subfigure}
        \begin{subfigure}{0.19\linewidth}
            \includegraphics[scale=0.95]{new_media/rama_tarflow_al6_2.pdf}
        \end{subfigure}
        \begin{subfigure}{0.19\linewidth}
            \includegraphics[scale=0.95]{new_media/rama_tarflow_al6_3.pdf}
        \end{subfigure}
        \begin{subfigure}{0.19\linewidth}
            \includegraphics[scale=0.95]{new_media/rama_tarflow_al6_4.pdf}
        \end{subfigure}
    \caption{\nameshort}
    \end{subfigure}
    \caption{Hexa-alanine Ramachandran plots for ECNF++ (SNIS) and \nameshort (SMC). \scione{4} points sampled per method.}
    \label{fig:ramas_al6}
\end{figure}

\subsection{Ablation Studies}
\label{sec:ablations}

\xhdr{Ablation of SNIS and SMC} In \cref{tab:smc_abalate_small,tab:smc_abalate_large} we compare the quantitative performance of ECNF++ and \nameshort as proposals, with SNIS, and in the case of \nameshort, with SMC. This table also presents the \ticametric results. We almost uniformly observe a significant reduction in \emetric using SNIS or SMC over the raw proposals, with only ECNF++ SNIS alanine tetrapeptide failing to achieve this. On alanine dipeptide we also observe a large decrease in macrostructure metrics \torusmetric and \ticametric, which is expected as the training data and subsequently the proposal distribution was intentionally biased to provide improved coverage \citep{klein2024transferable}. On all other datasets (and both models) we generally see no change or an increase in these metrics after reweighting with either SNIS or SMC, suggesting there is some tradeoff between matching the energy distribution whilst maintaining good macrostructure metrics. We lastly note that the use of SMC over SNIS does not uniformly improve performance on \emetric for \nameshort, with only alanine dipeptide and trialanine exhibiting this trend. However, we believe this may be an artifact of the good proposal overlap, and draw attention to the significant reduction in \emetric achieved on chignolin by SMC over SNIS in \cref{fig:chignolin_tarflow}.

\begin{table}[H]
    \centering
    \caption{Comparison of proposal, SNIS, and SMC for \nameshort and ECNF++ on alanine dipeptide and trialanine.}
    \label{tab:smc_abalate_small}
    \small
    \begin{tabular}{@{}lccccccc}
    \toprule
    Datasets $\rightarrow$ & \multicolumn3c{Alanine dipeptide} & \multicolumn3c{Trialanine}  \\
    \cmidrule(lr){2-4}\cmidrule(lr){5-7}
    Algorithm $\downarrow$ & \emetric & \torusmetric & \ticametric & \emetric & \torusmetric & \ticametric \\
    \midrule
    ECNF++ Proposal & 6.675 $\pm$ 0.297&	1.776 $\pm$ 0.018&	3.920 $\pm$ 0.025& 5.424 $\pm$ 1.595&	0.277 $\pm$ 0.004&	0.435 $\pm$ 0.009 \\
    ECNF++ SNIS & 0.914 $\pm$ 0.122&	0.189 $\pm$ 0.019&	0.402 $\pm$ 0.002&	2.206 $\pm$ 0.813&	0.962 $\pm$ 0.253&	0.597 $\pm$ 0.023 \\
    \midrule
    \nameshort Proposal & $\geq 10^4$ & 1.695 $\pm$ 0.015 & 3.862 $\pm$ 0.038 & $\geq 10^8$ & 0.338 $\pm$ 0.036 & 0.449 $\pm$ 0.028 \\
    \nameshort SNIS & 0.873 $\pm$ 0.338 & 0.439 $\pm$ 0.129 & 0.942 $\pm$ 0.268 &  0.758 $\pm$ 0.506 & 0.502 $\pm$ 0.016 & 0.518 $\pm$ 0.032 \\
    \nameshort AIS & 0.960 $\pm$ 0.617 & 0.430 $\pm$ 0.034 & 0.806 $\pm$ 0.166 &  0.754 $\pm$ 0.230 & 0.495 $\pm$ 0.033 & 0.476 $\pm$ 0.048 \\
    \nameshort SMC & 0.741 $\pm$ 0.189 & 0.431 $\pm$ 0.141 & 0.915 $\pm$ 0.316 &  0.598 $\pm$ 0.084 & 0.503 $\pm$ 0.029 & 0.501 $\pm$ 0.031 \\
    \bottomrule
    \end{tabular}
\end{table}

\begin{table}[H]
    \centering
    \caption{Comparison of proposal, SNIS, and SMC for \nameshort and ECNF++ on alanine tetrapeptide and hexa-alanine.}
    \label{tab:smc_abalate_large}
    \small
    \begin{tabular}{@{}lccccccc}
    \toprule
    Datasets $\rightarrow$ & \multicolumn3c{Alanine tetrapeptide} & \multicolumn3c{Hexa-alanine}  \\
    \cmidrule(lr){2-4}\cmidrule(lr){5-7}
    Algorithm $\downarrow$ & \emetric & \torusmetric & \ticametric & \emetric & \torusmetric & \ticametric \\
    \midrule
    ENCF++ Proposal & 2.983 $\pm$ 1.266&	0.576 $\pm$ 0.002&	0.737 $\pm$ 0.013 & $\geq 10^4$ &	1.136 $\pm$ 0.030&	0.688 $\pm$ 0.066 \\
    ECNF++ SNIS & 5.638 $\pm$ 0.483&	1.002 $\pm$ 0.061&	0.832 $\pm$ 0.021& 10.668 $\pm$ 0.285&	1.902 $\pm$ 0.055&	0.632 $\pm$ 0.087 \\ 
    \midrule
    \nameshort Proposal & $\geq 10^6$ & 0.624 $\pm$ 0.023 & 0.791 $\pm$ 0.050 & $\geq 10^{12}$ & 1.079 $\pm$ 0.153 & 0.299 $\pm$ 0.039 \\
    \nameshort SNIS & 1.068 $\pm$ 0.495 & 0.969 $\pm$ 0.067 & 0.879 $\pm$ 0.047 & 1.036 $\pm$ 0.534 & 1.473 $\pm$ 0.114 & 0.452 $\pm$ 0.245 \\
    \nameshort AIS & 1.070 $\pm$ 0.272 & 0.923 $\pm$ 0.100 & 0.920 $\pm$ 0.028 & 1.131 $\pm$ 0.384 & 1.510 $\pm$ 0.113 & 0.492 $\pm$ 0.240 \\
    \nameshort SMC & 1.007 $\pm$ 0.382 & 1.039 $\pm$ 0.069 & 0.904 $\pm$ 0.054 &  1.155 $\pm$ 0.635 & 1.517 $\pm$ 0.118 & 0.530 $\pm$ 0.198 \\
    \bottomrule
    \end{tabular}
\end{table}

\xhdr{Center of mass adjusted energy} As discussed in \S\ref{sec:scaling_training}, the \nameshort proposal distribution is not mean-free due to the CoM data augmentation applied to the training data, with a centroid norm distribution \(||C|| \sim \sigma \chi_3\). This can introduce adverse behavior, as the target energy is invariant to \(||C||\), and thus the importance weights will depend on \(||C||\) of a sample \(x\). Concretely, a low (target) energy sample generated far from the origin (with large \(||C||\)) will have low likelihood under \(p_\theta\) but high likelihood under \(p\) leading to a very large importance weight.

To provide a visual intuition for this effect, we plot in \cref{fig:com_logit_clipping} the centroid norm distribution of the \nameshort proposal samples before and after SNIS. Here the empirical distribution is generated using \(2\times10^7\) samples, to approximate the asymptotic behavior. In \cref{fig:com_0_0} we observe that, even with this extremely large number of samples, without weight clipping or center of mass adjusted energy \cref{eqn:com_adjust} the \(||C||\) distribution is greatly influenced by the reweighting. In this case resampling shifts most density to high \(||C||\) samples, where there was very little density prior to reweighting and hence little sample diversity, and a large peak manifests resulting from a single sample with very large importance weight. In contrast, in \cref{fig:com_1_0} we see a much smaller change in \(||C||\) distribution after reweighting, with no large peak for any given sample, after applying the CoM adjustment. In this case there is no overweighting of high \(||C||\) regions with limited sample diversity. Adding weight clipping to the standard proposal energy function (\cref{fig:com_0_1}) greatly reduces the change in distribution from reweighting, although the mean remains notably shifted towards higher \(||C||\) samples. Applying weight clipping with the CoM adjusted proposal energy function (\cref{fig:com_1_1}) has little effect on the \(||C||\) distribution beyond smoothing. We emphasize that these plots are presented for \(2 \times 10^7\) samples hence clipping may have a larger still effect for both proposal energy functions for smaller sample sets.

In Figures \ref{fig:com_augmentation_samples} and \ref{fig:com_augmentation_timesteps} we ablate the utility of performing the center of mass energy adjustment to the proposal energy. Specifically, we ablate the center of mass adjustment as a function of number of samples used during inference and also as a function of a number of inference time steps. Each of these ablations is performed on trialanine. Considering Figure \ref{fig:com_augmentation_samples}, there is little distinction between variants on \emetric with respect to \(N\) samples, although standard energy without clipping can be identified as the least performant, and all methods improve with increased \(N\). On $\sT$-$\gW_2$ the standard energy without clipping is again evidently the worst performing, with a clear benefit to applying the CoM adjustment where clipping is not used. The best performing variants do employ clipping, with a slight but clear benefit to using the center of mass adjustment. Considering Figure \ref{fig:com_augmentation_timesteps}, we observe again that the center of mass adjustment improves performance greatly where clipping is not employed, and notably still without clipping on \torusmetric.

\begin{figure}[t]
    \centering
      \begin{subfigure}{0.45\linewidth}
         \centering
         \includegraphics{new_media/coms_use_com_0_clip_logits_0.pdf}
        \caption{Weight clipping \xmark, CoM energy \xmark}
         \label{fig:com_0_0}
    \end{subfigure}
    \begin{subfigure}{0.45\linewidth}
        \centering
         \includegraphics{new_media/coms_use_com_1_clip_logits_0.pdf}
        \caption{Weight clipping \xmark, CoM energy \cmark}
        \label{fig:com_1_0}
    \end{subfigure}
    \begin{subfigure}{0.45\linewidth}
        \vspace{1em}
         \centering
         \includegraphics{new_media/coms_use_com_0_clip_logits_1.pdf}
        \caption{Weight clipping \cmark, CoM energy \xmark}
         \label{fig:com_0_1}
    \end{subfigure}
    \begin{subfigure}{0.45\linewidth}
        \vspace{1em}
        \centering
         \includegraphics{new_media/coms_use_com_1_clip_logits_1.pdf}
        \caption{Weight clipping \cmark, CoM energy \cmark}
        \label{fig:com_1_1}
    \end{subfigure}
    \caption{Centroid norm $||C||$ histograms for \(2\times10^7\) proposal samples and reweighted proposal samples, with / without both of weight clipping (0.2\%) and center of mass adjusted energy.}
    \label{fig:com_logit_clipping}
\end{figure}

\begin{figure}[t]
    \captionsetup[subfigure]{aboveskip=-1pt,belowskip=-1pt}
    \centering
      \begin{subfigure}{0.45\linewidth}
         \centering
         \includegraphics{new_media/energy_vs_samples_appendix.pdf}
         \label{fig:com_energy_vs_samples}
    \end{subfigure}
    \begin{subfigure}{0.45\linewidth}
        \centering
         \includegraphics{new_media/torus_vs_samples_appendix.pdf}
        \label{fig:com_torus_vs_samples}
    \end{subfigure}
    \caption{Trialanine \nameshort SMC \emetric and \torusmetric performance with standard and center of mass adjusted energy, with / without weight clipping (0.2\%) at a variety of sampling set sizes. 100 Langevin timesteps used.}
    \label{fig:com_augmentation_samples}
\end{figure}

\begin{figure}[h!]
    \vspace{10pt}
    \captionsetup[subfigure]{aboveskip=-1pt,belowskip=-1pt}
    \centering
      \begin{subfigure}{0.45\linewidth}
         \centering
         \includegraphics{new_media/energy_vs_timestep_appendix.pdf}
         \label{fig:com_energy_vs_samples}
    \end{subfigure}
    \begin{subfigure}{0.45\linewidth}
        \centering
         \includegraphics{new_media/torus_vs_timestep_appendix.pdf}
        \label{fig:com_torus_vs_samples}
    \end{subfigure}
    \caption{Trialanine \nameshort SMC \emetric and \torusmetric performance with standard and center of mass adjusted energy, with / without weight clipping (0.2\%) at a varierty of Langevin time discretizations. \scione{4} samples generated.}
    \label{fig:com_augmentation_timesteps}
\end{figure}

\clearpage
\xhdr{Ablation on EACF importance weight clipping}
We lastly report additional EACF results on alanine dipeptide. 
In our main results we use the same $0.2\%$ clipping threshold on the importance weights as other models for fair comparison. Nevertheless, in the resampling process, we observe a significant degradation in sample diversity, as evidenced by the energy histogram in \cref{fig:al2_energy_figs} and Ramachandran plots \cref{fig:rama-al23}. In \cref{fig:al2_eacf_rama_diff_clip} we plot energy histograms and Ramachandran plots for a variety of different clipping thresholds. We observe that EACF generates highly unreliable importance weights, particularly visible in the energy histograms where there are extreme spikes and poor alignment with the true data distribution. This leads to poor resampling quality, as demonstrated in the corresponding Ramachandran plots where the resampled points fail to capture the true data distribution. While increasing the clipping threshold to $10\%$ shows some improvement, the fundamental issue of inaccurate importance weight estimation by EACF persists across different clipping ratios.

\begin{figure}[H]
    \captionsetup[subfigure]{aboveskip=-1pt,belowskip=-1pt}
    \centering   
    \begin{subfigure}{0.33\linewidth}
        \centering
        \includegraphics[]{new_media/energies_eacf_al2_0.002.pdf}
    \end{subfigure}
    \begin{subfigure}{0.33\linewidth}
        \centering
        \includegraphics[]{new_media/energies_eacf_al2_0.02.pdf}
    \end{subfigure}
    \begin{subfigure}{0.33\linewidth}
        \centering
        \includegraphics[]{new_media/energies_eacf_al2_0.1.pdf}
    \end{subfigure}
    \begin{subfigure}{0.33\linewidth}
        \centering
        \includegraphics[]{new_media/rama_eacf_al2_0.002_0.pdf}
        \caption{Clip $0.2\%$}
    \end{subfigure}
    \begin{subfigure}{0.33\linewidth}
        \centering
        \includegraphics[]{new_media/rama_eacf_al2_0.02_0.pdf}
        \caption{Clip $2\%$}
    \end{subfigure}
    \begin{subfigure}{0.33\linewidth}
        \centering
        \includegraphics[]{new_media/rama_eacf_al2_0.1_0.pdf}
        \caption{Clip $10\%$}
    \end{subfigure}
    \caption{Energy histogram and Ramachandran Plots of $\sethree$-EACF under different weight clipping ratio $[0.2\%, 2\%, 10\%]$.}
    \label{fig:al2_eacf_rama_diff_clip}
\end{figure}

\end{document}